\documentclass{article}

\usepackage{microtype}
\usepackage{graphicx}
\usepackage{booktabs} %

\usepackage{hyperref}

\usepackage{algorithmic,algorithm}

\usepackage{amsmath}
\usepackage{amssymb}
\usepackage{mathtools}
\usepackage{amsthm}
\usepackage{enumitem}
\usepackage[capitalize,noabbrev]{cleveref}

\theoremstyle{plain}

\theoremstyle{definition}

\theoremstyle{remark}

\usepackage[textsize=tiny]{todonotes}

\usepackage{color}
\usepackage{subcaption}
\usepackage{diagbox}
\usepackage{caption}
\usepackage{wrapfig}

\newcommand{\methodname}{TALAR}
\newcommand{\tl}{L_{\rm T}}
\newcommand{\nl}{L_{\rm N}}
\newcommand{\expect}{\mathbb{E}}

\def\gA{{\mathcal{A}}}

\def\gD{{\mathcal{D}}}

\def\gM{{\mathcal{M}}}

\def\gS{{\mathcal{S}}}

\newtheorem{defn}{Definition}

\usepackage[in]{fullpage}  
\usepackage{natbib} 
\usepackage{mathpazo} 

\usepackage[utf8]{inputenc} %
\usepackage[T1]{fontenc}    %
\usepackage{hyperref}       %
\usepackage{url}            %
\usepackage{booktabs}       %
\usepackage{amsfonts}       %
\usepackage{nicefrac}       %
\usepackage{microtype}      %
\usepackage{lipsum}         %
\usepackage{todonotes}

\hypersetup{
    colorlinks=true,
    citecolor=orange,
    linkcolor=blue
}

\AtBeginDocument{\setlength\abovedisplayskip{5pt}}
\AtBeginDocument{\setlength\belowdisplayskip{5pt}}

\usepackage{tabulary}

\usepackage{amssymb}%
\usepackage{pifont}%

\usepackage{url}            %
\usepackage{booktabs}       %
\usepackage{multirow}    
\usepackage{amsfonts}       %
\usepackage{nicefrac}       %
\usepackage{microtype}      %
\usepackage{natbib}
\usepackage{enumerate}
\usepackage{hhline}
\usepackage{makecell}
\usepackage{pifont}

\usepackage{graphicx} %
\usepackage{caption}
\usepackage{subcaption}
\usepackage{amsmath}
\usepackage{amsthm}
\usepackage{amssymb}
\usepackage{tikz}
\usepackage{xcolor}
\usetikzlibrary{arrows}

\allowdisplaybreaks

\usepackage{mathrsfs}

\usepackage{hyperref}
\usepackage{bm}

\renewcommand{\cite}{\citep}

\title{Natural Language-conditioned Reinforcement Learning with Inside-out Task Language Development and Translation}
\date{\vspace{-5ex}}

\usepackage{authblk}

\author[1,2]{Jing-Cheng Pang \thanks{Equal contribution. Email: \texttt{pangjc@lamda.nju.edu.cn} and \texttt{yangxy@lamda.nju.edu.cn}}}
\author[1,2]{Xin-Yu Yang{ $^*$}}
\author[1,2]{Si-Hang Yang}
\author[1,2]{Yang Yu\thanks{Corresponding author. Email: \texttt{yuy@nju.edu.cn}}}
\affil[1]{National Key Laboratory for Novel Software Technology, Nanjing University}
\affil[2]{Polixir Technologies}

\begin{document}

\maketitle
\begin{abstract}
    Natural Language-conditioned reinforcement learning (RL) enables the agents to follow human instructions. Previous approaches generally implemented language-conditioned RL by providing human instructions in natural language (NL) and training a following policy. In this \emph{outside-in} approach, the policy needs to comprehend the NL and manage the task simultaneously. However, the unbounded NL examples often bring much extra complexity for solving concrete RL tasks, which can distract policy learning from completing the task. To ease the learning burden of the policy, we investigate an \emph{inside-out} scheme for natural language-conditioned RL by developing a task language (TL) that is task-related and unique. The TL is used in RL to achieve highly efficient and effective policy training. Besides, a translator is trained to translate NL into TL. We implement this scheme as \methodname~(\textbf{TA}sk \textbf{L}anguage with predic\textbf{A}te \textbf{R}epresentation) that learns multiple predicates to model object relationships as the TL. Experiments indicate that \methodname~not only better comprehends NL instructions but also leads to a better instruction-following policy that improves 13.4\% success rate and adapts to unseen expressions of NL instruction. The TL can also be an effective task abstraction, naturally compatible with hierarchical RL.
\end{abstract}

\section{Introduction}
\begin{figure}[t]
\includegraphics[width=1\linewidth]{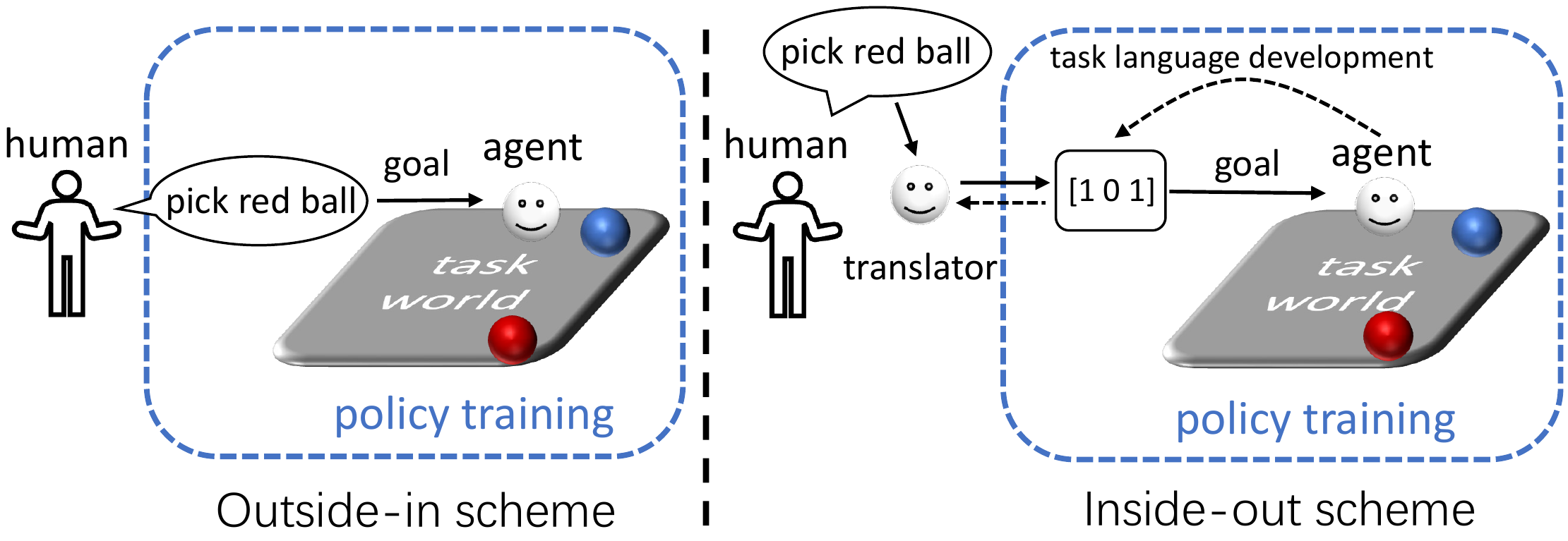}
\caption{An illustration of OIL and IOL schemes in NLC-RL.
\textbf{Left:} OIL directly exposes the NL instructions to the policy. \textbf{Right:} IOL develops a task language, which is task-related and a unique representation of NL instructions. The solid lines represent instruction following process, while the dashed lines represent TL development and translation.}
\label{fig:learning_manner}
\end{figure}

Enabling the robot to work with humans is a hallmark of machine intelligence. Language is a vital connection between humans and machines \cite{hallmark_of_machine_intell}, and it has been investigated for instructing robot execution, designing rewards, and serving as observation or action in reinforcement learning (RL) \cite{survey_of_RL_by_NL}. 
We are especially interested in developing agents that follow human instructions in this broad context. Natural language-conditioned reinforcement learning (NLC-RL) is a promising tool in this pursuit, which provides the policy with human instructions in natural language (NL) and trains the policy with RL algorithms. 
In this \emph{outside-in} learning (OIL, Figure \ref{fig:learning_manner}-left) scheme, the policy is directly exposed to the NL instructions. Thus, the policy must comprehend the NL instructions and complete the RL tasks simultaneously.

However, natural language is an unbounded representation of human instruction, which imposes an additional burden on the policy when solving concrete RL tasks. For example, to ask a robot to bring a drink, one may say: \textit{Get me a drink}, while another may ask: \textit{Can you take the beverage to me?} Despite having identical meanings, the two NL instructions are expressed in vastly different ways. To complete human instructions, the policy must simultaneously comprehend these diverse, unbounded NL instructions and solve the RL tasks, resulting in inefficient policy learning.

In this paper, we investigate an \textit{Inside-Out} Learning (IOL) scheme to enable efficient and effective policy learning in NLC-RL, as depicted in Figure \ref{fig:learning_manner}-right. 
The IOL develops a task language (TL) that is task-related and uniquely represents human instruction. The TL can be utilized in RL to alleviate the burden of policy learning by comprehending diverse NL expressions. 
In addition to developing TL, IOL trains a translator that translates NL into TL. 
A crucial aspect of IOL is how the task language is represented. We believe that \emph{expressiveness} and \emph{conciseness} are essential properties of language representation. Expressiveness ensures that the TL accurately reflects human instruction, while conciseness facilitates policy comprehension.
To satisfy these two requirements, we propose representing TL with predicate representation, which is deemed expressive \cite{expressive_power_of_FOL} and concise as a discrete representation.

We introduce an implementation of IOL, called \methodname~for \textbf{TA}sk \textbf{L}anguage with predic\textbf{A}te \textbf{R}epresentation. \methodname~consists of three components: (1) a TL generator that generates TL with predicate representation, (2) a translator that translates NL into TL and (3) a policy that solves the RL tasks assigned by human instructions. Specifically, the TL generator develops TL through the identification of object relationships. To achieve this, the TL generator learns multiple (anonymous) predicates and their arguments to model the relationships between objects. The translator is trained to translate NL into TL using the variational auto-encoder \cite{vae} algorithm. With the optimized translator, \methodname~trains an instruction-following policy that completes human instructions.

Our contributions include the following: We propose a novel NLC-RL scheme, IOL, that enables highly efficient policy learning. IOL develops TL that serves as a unique representation of human instruction and trains the policy following TL. 
Besides, we present a specific IOL implementation, including a neural network architecture that automatically discovers relationships between the objects.
Through our experiments in the CLEVR-Robot environment \cite{language_as_abstraction}, we find that \methodname~can better translate diverse NL expressions into a unique representation compared to traditional OIL methods. A policy can learn to complete human instructions efficiently and adapt to unseen NL instructions with the resulting TL. Moreover, the resulting TL effectively represents human instruction, providing a solid baseline for abstraction in hierarchical RL \cite{hiecarchical_rl}.

\section{Related Work}
\label{sec:related_works}

This section begins with a summary of prior research on instruction following with RL, followed by two paragraphs discussing works pertinent to our methodology, i.e., language generation in RL and language translation.

\textbf{Instruction following with RL.}
Instruction-following problems require agents to perform tasks specified by NL instructions. Previous methods employ RL to train an instruction-following policy and expose the NL directly to the policy. For example, 
\cite{instruction_follow_4} encodes a single-sentence instruction with a pre-trained language model and feeds the policy with the NL encoding. 
\cite{instruction_follow_2} learns a policy that maps NL instructions to action sequences by marginalizing implied goal locations. 
\cite{instruction_follow_1} combines human instructions with agent observations via a multiplication-based mechanism and then pre-trains the instruction-following policy using behaviour cloning \cite{BC}.
Instead of exposing NL to the policy, \cite{binary_language_repres} encodes NL to a manually-designed binary vector in which each element has meaning. 
Besides, the instruction-following policy has close ties to Hierarchical RL \cite{hiecarchical_rl} because the instructions can be naturally viewed as a task abstraction for a low-level policy \cite{hrl_example}.
HAL \cite{language_as_abstraction} takes advantage of the compositional structure of NL and makes decisions directly at the NL level to solve long-term, complex RL tasks. 
These previous methods either expose the unbounded NL instructions directly to the policy or encode the NL instructions to a scenario-specific manual vector, both of which have limitations. In contrast to them, we propose developing a task-related task language that is a unique representation of NL instruction and is, therefore, easily understood by the policy.

\textbf{Language representation in RL.} We are interested in language representation, which is fundamental to developing task language. RL communities often consider language representation when agents learn effective message protocol to communicate with their partners \cite{emergent_com_1,emergent_com_2,emergent_com_3,one_hot_discrete_emergent_language}. 
Motivated by the discrete nature of human language, discrete representation has been widely used in prior research. 
For example, \cite{discrete_communication} enables agents to communicate using discrete messages and demonstrates that discrete representation has comparable performance to continuous representation with a much smaller vocabulary size. 
One-hot representation \cite{one_hot_discrete_emergent_language,one_hot_sender_receiver} and binary representation \cite{binary_language_repres,dec_POMDP} are prevalent forms of discrete language representation. For instance, 
\cite{one_hot_discrete_emergent_language} uses one-hot language representation to allow two agents to communicate to differentiate between images. 
However, these representations only employ discrete symbols and operate on a propositional level, lacking the ontological commitment of the predicate representation that the world consists of objects and their relationships \cite{artificial_intel_intro}. 
In this paper, we develop the task language following the discrete form of language representation while the predication representation is used.

\textbf{Language translation.} In this paper, \methodname~translates NL into TL, which lies in the domain of machine translation \cite{machine_translation}. 
In this field of study, numerous approaches have been developed \cite{machine_translation_survey,machine_translation_survey_2}. Encoder-decoder \cite{encoder_decoder} is a promising machine translation tool because of its ability to extract the effective features of the input sentences. For example, \cite{translation_vae} proposes to utilize a continuous latent variable as an efficient machine translation feature based on a variational auto-encoder \cite{vae}. In this paper, we adhere to this class of machine translation techniques that employ an encoder-decoder structure, treating the NL as the source language and the TL as the target language.

\section{Background}
\label{sec:background}

\textbf{RL and NLC-RL.} A typical RL task can be formulated as a Markov Decision Process (MDP) \cite{puterman2014markov, sutton2011reinforcement}, which is described as a tuple $\gM = \left( \gS, \gA, P, r, \gamma, d_0  \right)$. Here $\gS$ represents the state space. $\gA$ is the finite action space defined by $\gA = \{a_0, a_1, \cdots, a_{\vert \gA \vert - 1} \}$. $P$ represents the probability of transition while $r$ represents the reward function. $\gamma$ is the discount factor determining the weights of future rewards, whereas $d_0$ is the initial state distribution. A policy $\pi: \gS \rightarrow \Delta (\gA)$ is a mapping from state space to the probability space over action space. 
In NLC-RL, the agent receives an NL instruction ($L$) that reflects the human's instruction on the agent. An instruction-following policy $\pi(\cdot|s_t, L)$ is trained to make decisions based on the current state $s_t$ and NL instruction $L$. 
The overall objective of NLC-RL is to maximize the expected return under different NL instructions:
\begin{equation}
    J = \expect \bigg[ \sum_{t=0}^\infty \gamma^t r(s_t,a_t|L) \big| s_0 \sim P, a_t \sim \pi (\cdot|s_t, L)\bigg].
\end{equation}
For the accuracy of sake, we use $\nl$ and $\tl$ to denote NL and TL, respectively.

\textbf{Predicate representation} uses discrete binary vectors to represent the relationship between objects or the property of an individual object. For example, the predicate representation vector [1, \textcolor{red}{1, 0, 0}, \textcolor{blue}{0, 1, 0}] could represent a predicate expression \texttt{Pred(\textcolor{red}{a},\textcolor{blue}{b})}. In this instance, \texttt{Pred} is a predicate that represents a relationship, and the symbols \texttt{a} and \texttt{b} are its arguments. 
In the vector, the first code [1] in the vector indicates that the value of \texttt{Pred} is True (i.e., the relationship holds), whereas the red and blue one-hot codes represent the indexes of arguments \texttt{a} and \texttt{b}, respectively.
It is demonstrated that predicate representation can be attained by employing networks to output predicate values. This way, the learning system can automatically identify relationships in a block stack task \cite{predicate_net_icaps}. In \methodname, neural networks learn both predicates and their arguments. Refer to Appendix \ref{appendix:discussion} for supplementary discussions on predicate representation.

\textbf{Natural language as input for the neural network.} NL sentences are in variable lengths and cannot be fed directly into a fully connected network. A standard solution is to encode each word as an embedding \cite{word2vec} and loop over each embedding using a recurrent model.
Except for the recurrent model, Bert \cite{bert} tokenizes the words before extracting sentence embedding features based on these tokens. Transformer \cite{transformer} is the predominant model for natural language processing.  
Since we only need to convert NL sentences to fixed-length encoding in our experiments, we employ a pre-trained, lightweight Bert model.

\begin{figure}[t]
    \centering
    \includegraphics[width=1\linewidth]{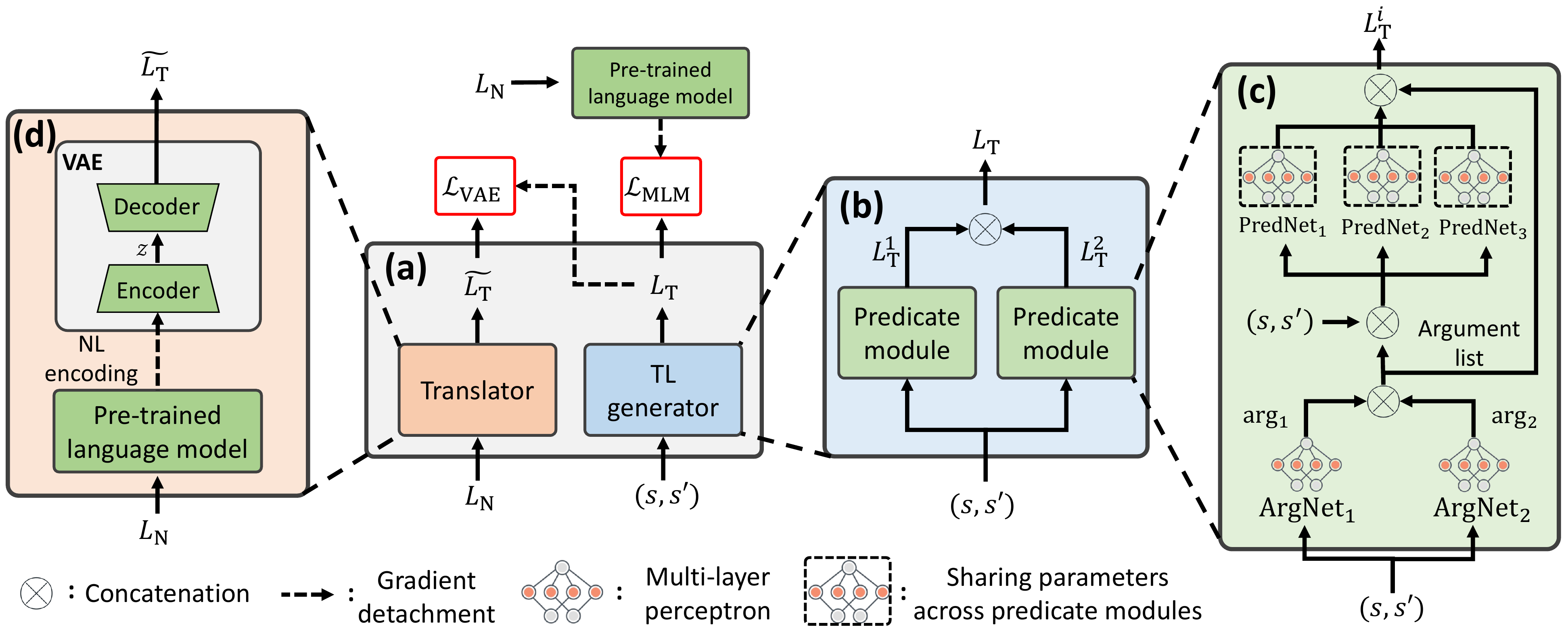}
    \caption{Overall training process of task language development and translation. \textbf{(a)} The overall training process. \textbf{(b)} Network architecture of the TL generator. \textbf{(c)} Architecture of one predicate module. \textbf{(d)} Network architecture of the translator.
    The number of predicate modules, arguments and predicate networks can be adjusted according to the task scale. 
    }
    \label{fig:overall_framework}
\end{figure}
    
\section{Method}
\label{sec:method}

This section presents our TALAR method for efficient policy learning in NLC-RL, which is based on the IOL scheme. We begin by introducing the task dataset for task language development and translation.

\begin{defn}[\textbf{Task dataset}]
\emph{
A task dataset $\gD=\{(s,s',\nl)_i\}$ consists of multiple triplets. Each triplet contains a natural language sentence $\nl$ and a task state pair $(s,s')$, where $\nl$ describes state change from $s$ to $s'$ in natural language.
}
\end{defn}

We use a state pair instead of a single state for the following reasons: (1) NL instruction frequently describes a state change, e.g., turn the wheel to the left; (2) it is not easy to describe a single state concisely in complex task scenarios. We let a person describe each state pair to collect the task dataset. 
Figure \ref{fig:overall_framework} illustrates how \methodname~makes use of task dataset for task language development and translation.
The subsequent subsections will elaborate on three critical components of \methodname: TL generation in predicate representation, NL translation by recovering TL, and policy training with reinforcement learning. Appendix \ref{appendix:algorithm} presents a summary of \methodname's training procedures.

\subsection{TL Generation in Predicate Representation}
\methodname~trains a TL generator $g_\theta(s,s')$ represented by neural networks and parameterized with $\theta$, which takes a state pair $(s,s')$ as the input and outputs task language $L_T$.
Next, we will introduce how task language is developed by describing the network structure of the TL generator and how to optimize the TL generator.

\textbf{Network architecture of TL generator.} As depicted in Figure \ref{fig:overall_framework}(b-c), the TL generator comprises $N_{\rm pm}$ instances of predicate modules (PM). Each PM first extracts $N_{\rm a}$ arguments $(\text{arg}_1,\text{arg}_2,\cdots,\text{arg}_{N_{\rm a}})$ according to the input state pair, and then determines the Boolean values of $N_\text{pn}$ predicates, given the extracted argument list. The predicate values are concatenated with the argument list and form the task language $\tl^i$ of the $i$-th module. Finally, the TL generator concatenates all PMs' output and generates the task language $\tl$.
Note that the number of PMs $N_{\rm pm}$, arguments $N_{\rm a}$, and predicates $N_{\rm pn}$ can be modified based on the RL task scale.

Specifically, each PM extracts the arguments through argument networks, denoted by ${\rm ArgNet}_i(s,s')$. An argument network is implemented as fully-connected networks ending with a Gumbel-Softmax activation layer \cite{gumbel_softmax}. Through the Gumbel-Softmax, the argument network is able to output a discrete one-hot vector $\text{arg}_i$ in form like $(0,1,\cdots,0)$, which represents an anonymous object. \methodname~utilizes multiple predicate networks, denoted by ${\rm PredNet}_i(s,s',\text{arg}_1,\cdots,\text{arg}_{N_{\rm a}})$, to determine the Boolean values of a set of anonymous predicates. Each predicate network outputs a 0-1 value, ending with a Gumbel-Softmax layer. All these 0-1 values are concatenated together with the argument list, yielding the task language $\tl^i$ of the $i$-th PM. 
Note that without the argument list contained in $\tl^i$, the resulting language cannot express different objects and therefore loses its expressiveness.

All predicate networks within the same PM receive the identical argument list. In the TL generator, there are multiple PMs, each possessing its argument networks, while the parameters of each predicate network $\text{PredNet}_i$ are shared across PMs. 
The parameter sharing here makes the $\textbf{PredNet}_i$ in different PMs identical, requiring them to capture consistent relations among the various arguments because they accept different arguments across PMs.
Finally, the total task language is represented by $\tl=[\tl^1,\cdots,\tl^{N_{\rm pm}}]$, which is a discrete binary vector. The Gumbel-Softmax activation technique permits the differentiation of the entire TL generation procedure.

\textbf{Training of the TL generator.} 
Training the TL generator ensures that the generated TL captures the crucial aspects of a given state transition. Based on this idea, \methodname~uses the Masked Language Modeling (MLM) technique \cite{bert} to train the TL generator. 
MLM masks $\nl$ sentences at random. 
For example, when the original sentence is \textit{It is a happy day}, the masked sentence could be \textit{It \texttt{Mask} a happy \texttt{Mask}}.
Then, MLM utilizes $\tl=g_\theta(s,s')$ and the masked $\nl$ to predict the masked words. 
We implement the above process using a pre-trained Bert language model (LM), which first tokenizes $\nl$ into tokens $T$. Then, \methodname~selects two random token positions of $T$ and replaces each with a unique token \texttt{Mask}. 
The TL generator is trained to optimize an MLM loss, which aims to predict the original tokens with masked tokens and task language:
\begin{equation}
\label{eq:mlm_loss}
    \mathcal{L}_{\rm MLM}(\theta) = \mathop{\expect}_{(s,s',\nl) \sim \gD} \left[ -\sum_{i \in M} \log f(T_i|b(T_{\setminus M}),g_\theta(s,s')) \right],
\end{equation}
where $M$ denotes the set of the masked positions, $T_{\setminus M}$ denotes the masked version of $\nl$'s tokens, $T_i$ is the $i$-th token, $b$ is the Bert model, and $f$ is a fully-connected network. Note that $f$ is also optimized via gradient backpropagation. We omit the notion of its parameters for simplicity.

\subsection{NL Translation by Recovering TL}
\label{sec:method_translator}
The objective of the translator is to translate the NL to the TL.
\methodname~trains the translator using variational auto-encoder (VAE) \cite{vae}. Specifically, given a TL $\tl=g_\theta(s,s')$ and corresponding NL $\nl$, we expect the VAE can recover $\tl$ from $\nl$. 
Figure \ref{fig:overall_framework}(d) presents the structure of the translator. 
\methodname~uses a pre-trained LM to process $\nl$ and a VAE to recover the task language. We let $\widetilde{\tl}$ denote the TL generated by translator, $q_{\phi_1}$ the VAE encoder parameterized with $\phi_1$, and $p_{\phi_2}$ the VAE decoder parameterized with $\phi_2$. Then the VAE is trained to minimize the VAE loss:
\begin{equation}
\label{vae_loss}
        L_{\rm VAE}(\phi_1,\phi_2) = \mathop{\expect}_{(s,s',\nl) \sim \gD}   \big[ D_{\rm KL}(q_{\phi_1}(z|L_{N}) , p(z))  -
         \mathop{\expect}_{z \sim q_{\phi_1}(\cdot|\nl)} \left[  \log p_{\phi_2}(\tl | z) \right]  \big],
\end{equation}
where $\tl=g_\theta(s,s')$, $z \sim q_{\phi_1}(\cdot|\nl)$ is the encoding generated by VAE encoder, and $D_{\rm KL}$ stands for KL-divergence.

We choose VAE because of its capacity to learn effective latent features. However, VAE is not essential to implement IOL, as the translator can be trained using alternative supervised learning methods. We demonstrate the effectiveness of VAE empirically in Section \ref{sec:exp_ablation}. While some may be concerned about the translator's ability to recover $\tl$ from $\nl$ accurately, we emphasize that the translator is primarily responsible for recovering the key positions that reflect the NL instructions and not the entire $L_T$. With the optimized translator, an instruction-following policy is trained to complete the human instructions, as described below.

\subsection{Policy Training With Reinforcement Learning}
\label{sec:method_policy_training}

\methodname~uses reinforcement learning to train an Instruction-Following Policy (IFP) $\pi(\cdot|s,\widetilde{\tl})$. When the agent collects samples from the environment, the task generates a random human instruction in NL, which is then translated into the task language $\widetilde{\tl}$ by the translator. Next, the IFP makes decisions for the entire episode based on the current observation and $\widetilde{\tl}$ until completing the instruction or reaching the maximum timestep. The IFP can be optimized with an arbitrary RL algorithm using the samples collected from the environments. In our implementation, we use PPO \cite{PPO} for \methodname~and all baselines. Note that during IFP training, the translator's parameters are fixed to prevent the translator from overfitting the current IFP.

\section{Experiments}
We conduct multiple experiments to evaluate \methodname~and answer the following questions: 
(1) Can \methodname~translate diverse NL instructions into a unique representation? (Section \ref{sec:exp_language_generation}) (2) How does \methodname~compare to traditional NLC-RL approaches in the instruction-following task? (Section \ref{sec:exp_ifp}) (3) Can TL serve as an abstraction for hierarchical RL? (Section \ref{sec:exp_hrl}) (4) How does every component influence the performance of \methodname? (Section \ref{sec:exp_ablation})

\begin{figure}[t]
\centering{
\includegraphics[width=0.6\textwidth]{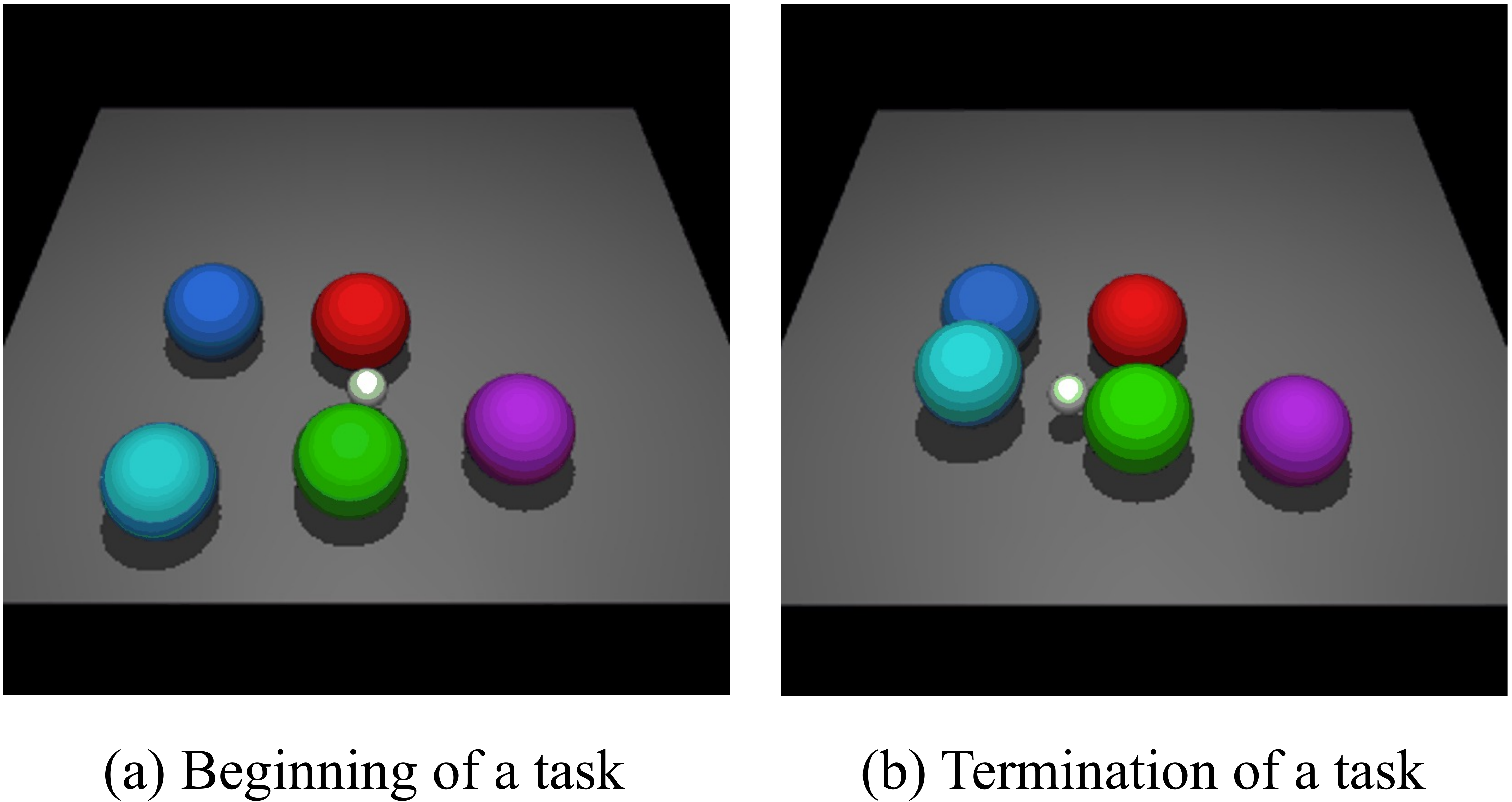}
}
\caption{A visualization of CLEVR-Robot environment in our experiments. (a) In the beginning, one NL instruction is randomly sampled as \textit{Can you move the \textcolor{cyan}{cyan} ball in front of the \textcolor{blue}{blue} ball?} Then agent executes actions to complete the instruction. (b) The task terminates if achieving the goal or reaching the maximum timestep.} 
\label{fig:env}
\end{figure}

We perform experiments in CLEVR-Robot environment \cite{language_as_abstraction}, as shown in Figure \ref{fig:env}. CLEVR-Robot is an environment for object interaction based on the MuJoCo physics engine \cite{mujoco}. The environment contains five movable balls and an agent (silverpoint). 
In each trajectory, the agent aims to complete a human instruction in NL that represents moving a specific ball to one direction (i.e., one of \texttt{[front, behind, left, right]}) of a destination ball. For example, an NL instruction can be \textit{Move the red ball to the left of the blue ball}, or \textit{Can you push the yellow ball to the right of the green ball?}
There are a total of $80$ distinct human instructions.
We use $18$ different NL sentence patterns for each human instruction to describe it, yielding 1440 different NL instructions.

To acquire the task dataset, we first train a policy that could move any specified ball to a specified position with PPO algorithms. Then, this policy will collect 100,000 state transitions, each corresponding to one random ball movement. Then, each state transition is assigned an NL description. 
We use Bert-base-uncased \cite{bert} as all pre-trained language models in our experiments. All experiments are performed with different random seeds five times, and the shaded area in the figures represents the standard deviation across all five trials. We refer readers to Appendix \ref{appendix:implmentation} for additional implementation details.

\subsection{Task Language Development and Translation}
\label{sec:exp_language_generation}
\begin{figure}[t]
\centering
\subcaptionbox{Task language\label{fig:exp_tl_tsne}}{
\includegraphics[width=0.25\textwidth]{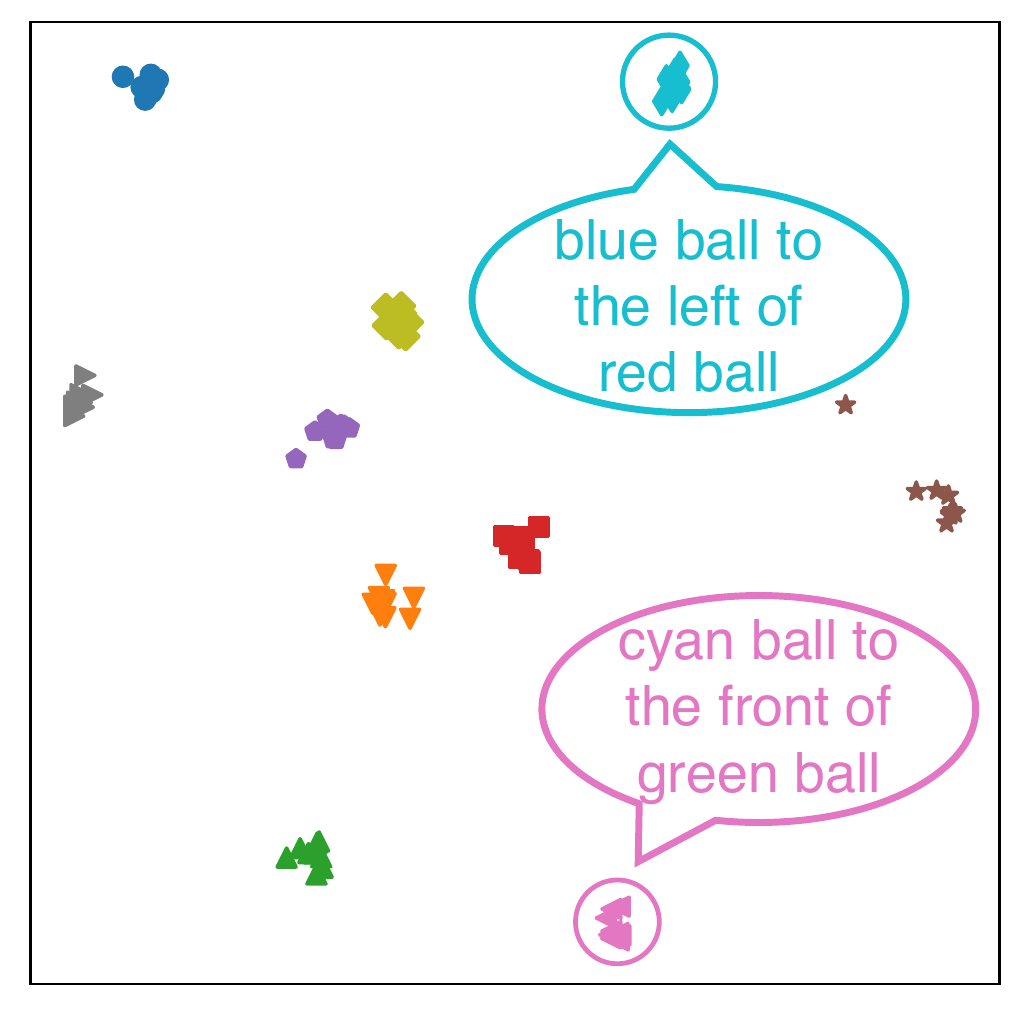}
}
\subcaptionbox{Bert\label{fig:exp_bert_tsne}}{
\includegraphics[width=0.25\textwidth]{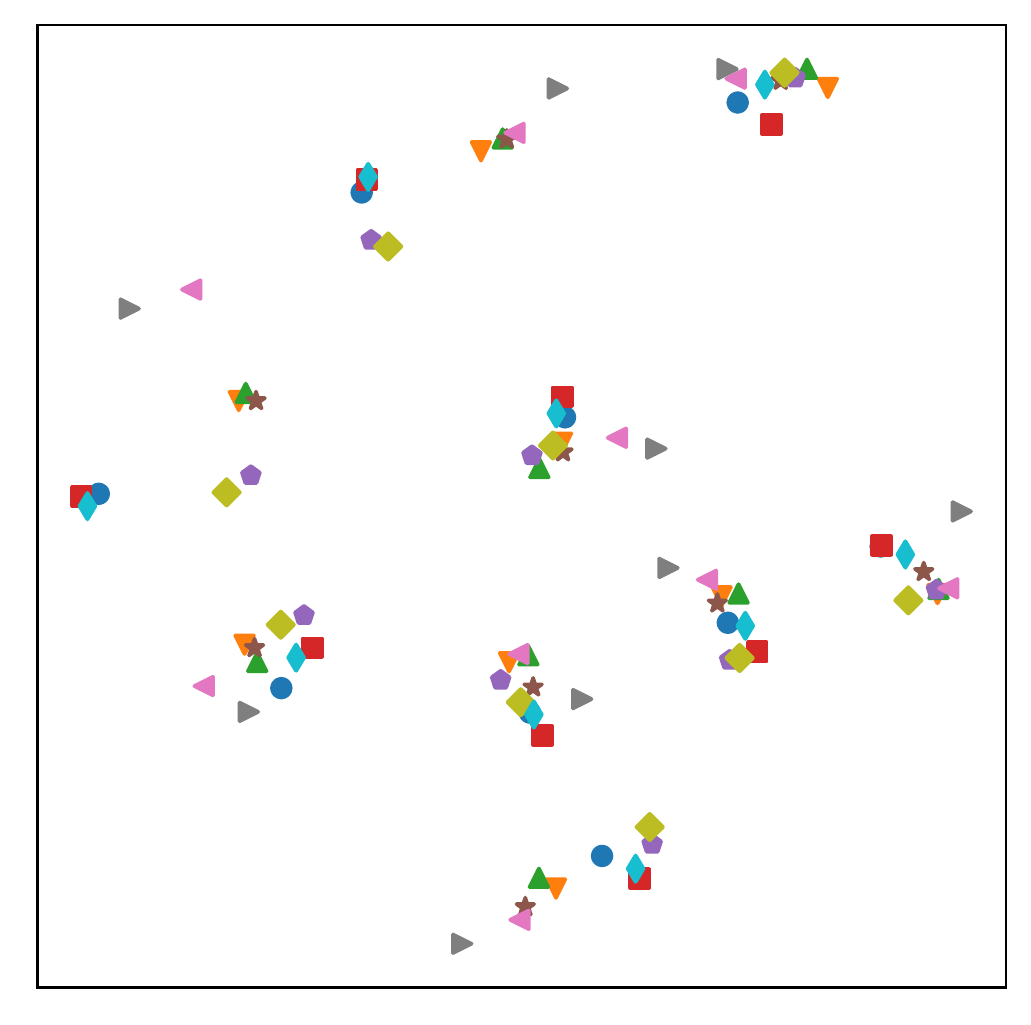}
}
\subcaptionbox{OIL Baseline\label{fig:exp_baseline_tsne}}{
\includegraphics[width=0.25\textwidth]{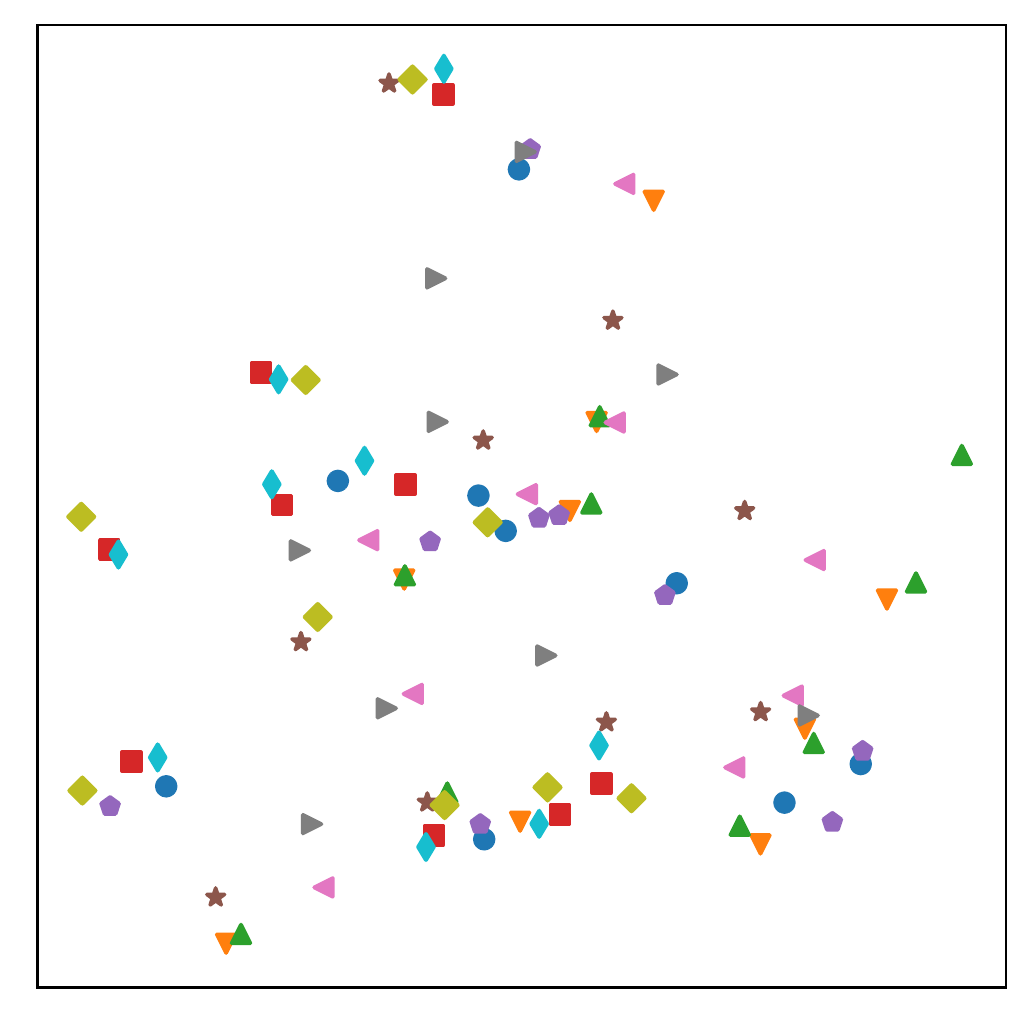}
}
\caption{The t-SNE representations of different types of NL encoding. Points with the same marker stand for the encoding of nine different NL expressions that describe the same human instruction. We add a slight noise to the overlapping points for better presentation. \textbf{(a)} The t-SNE representations of the TL output by the translator. \textbf{(b)} The encoding output by Bert model. \textbf{(c)} The encoding output by the language encoding layer of the OIL baseline (Bert-continuous in Section \ref{sec:exp_ifp}).} 
\label{fig:translator_result}
\end{figure}

We first verify whether the \methodname~can translate diverse NL expressions into a unique representation.
To answer the question, we randomly sample 10 different human instructions. Each instruction is expressed using nine NL sentence patterns for ninety NL sentences. Then, the optimized translator translates these NL sentences into TL. As depicted in Figure \ref{fig:exp_tl_tsne}, we project the resulting TL onto a two-dimensional plane using t-SNE \cite{tsne}.
Based on the projection results, we observe that \methodname~learns a unique representation of NL. This conclusion can be obtained because TL can represent different NL expressions for the same human instruction in a close area. 
As a comparison, we also project the NL encoding directly output by a pre-trained Bert model, as shown in Figure \ref{fig:exp_bert_tsne}. The points produced by Bert are scattered everywhere on the plane, indicating that a pre-trained Bert model fails to represent diverse human instructions uniquely. Besides, as depicted in Figure \ref{fig:exp_baseline_tsne}, an OIL baseline cannot distinguish the same human instruction with different NL expressions. This result suggests that the OIL baseline treats distinct NL expressions as distinct task objectives, which could slow policy learning. We refer readers to Appendix \ref{appendix:extra_exp_tsne} for more experiment results about the t-SNE projections.

\begin{figure}[htbp]
\centering{
\includegraphics[width=0.8\textwidth]{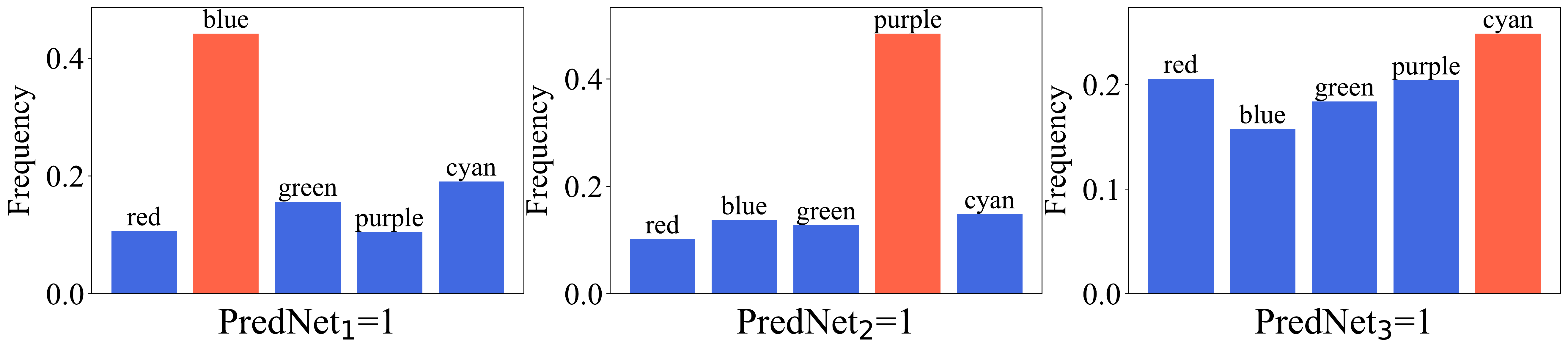}
}

\caption{Frequency of five destination balls when a predicate network outputs a value of $1$. Each bar stands for the frequency of the ball with a certain colour.} 
\label{fig:interpret}
\end{figure}

In addition to the above results, we observe that the generated TL is interpretable to some extend. We use the TL generator to output the TL regarding all state transitions in the task dataset, and observe the resulting TL and its the corresponding NL descriptions. Consequently, the output of the predicate network is related to the destination ball. Figure \ref{fig:interpret} presents the frequency of five destination balls when the predicate network outputs $1$. $\text{PredNet}_1$ and $\text{PredNet}_2$ clearly target the blue and purple balls, respectively. However, $\text{PredNet}_3$ is more difficult to interpret than the other two networks. There could be other relations other than with the destination ball.

\begin{figure}[t]
\centering
\subcaptionbox*{}{
\includegraphics[width=0.315\textwidth]{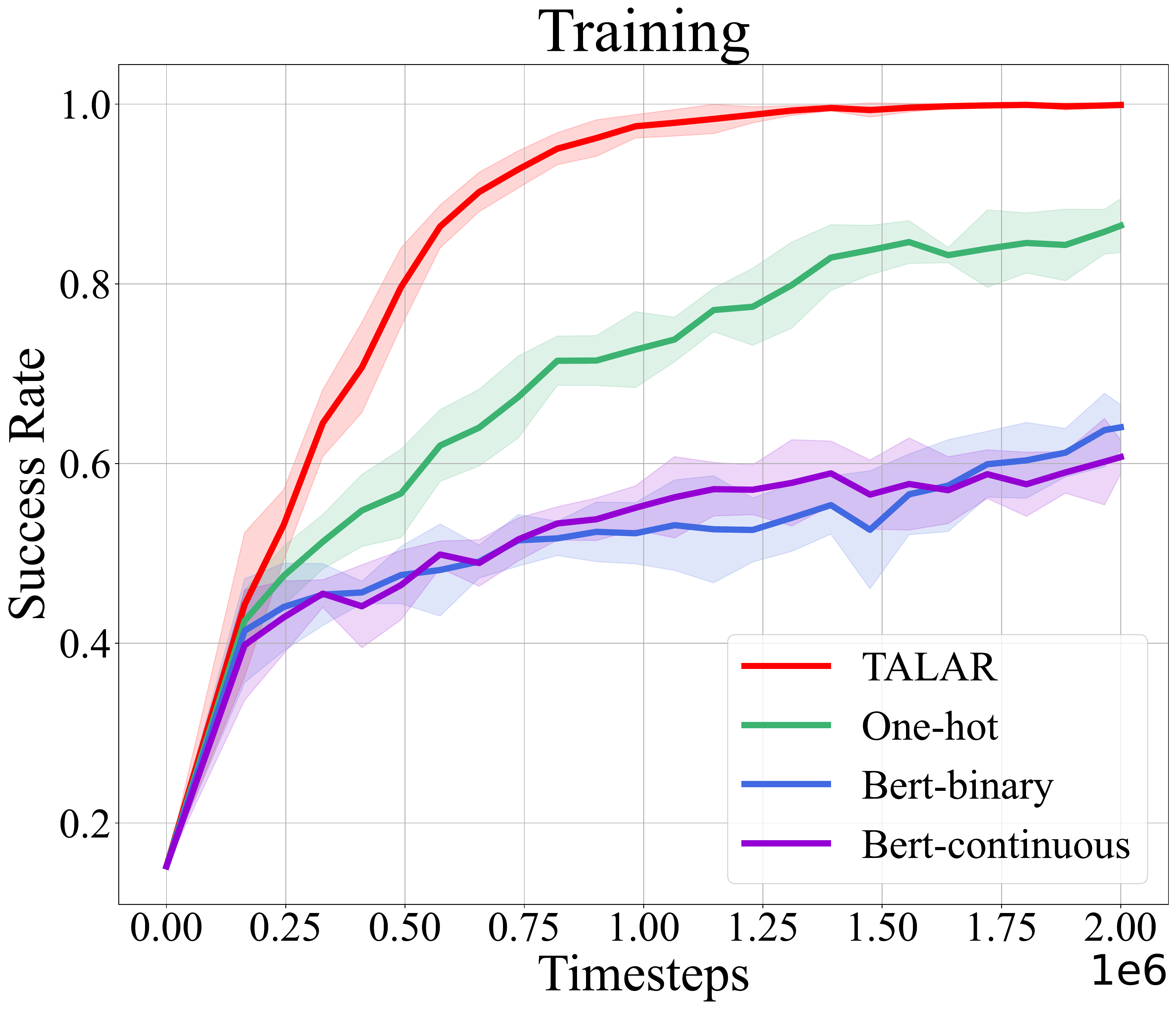}
}
\subcaptionbox*{}{
\includegraphics[width=0.315\textwidth]{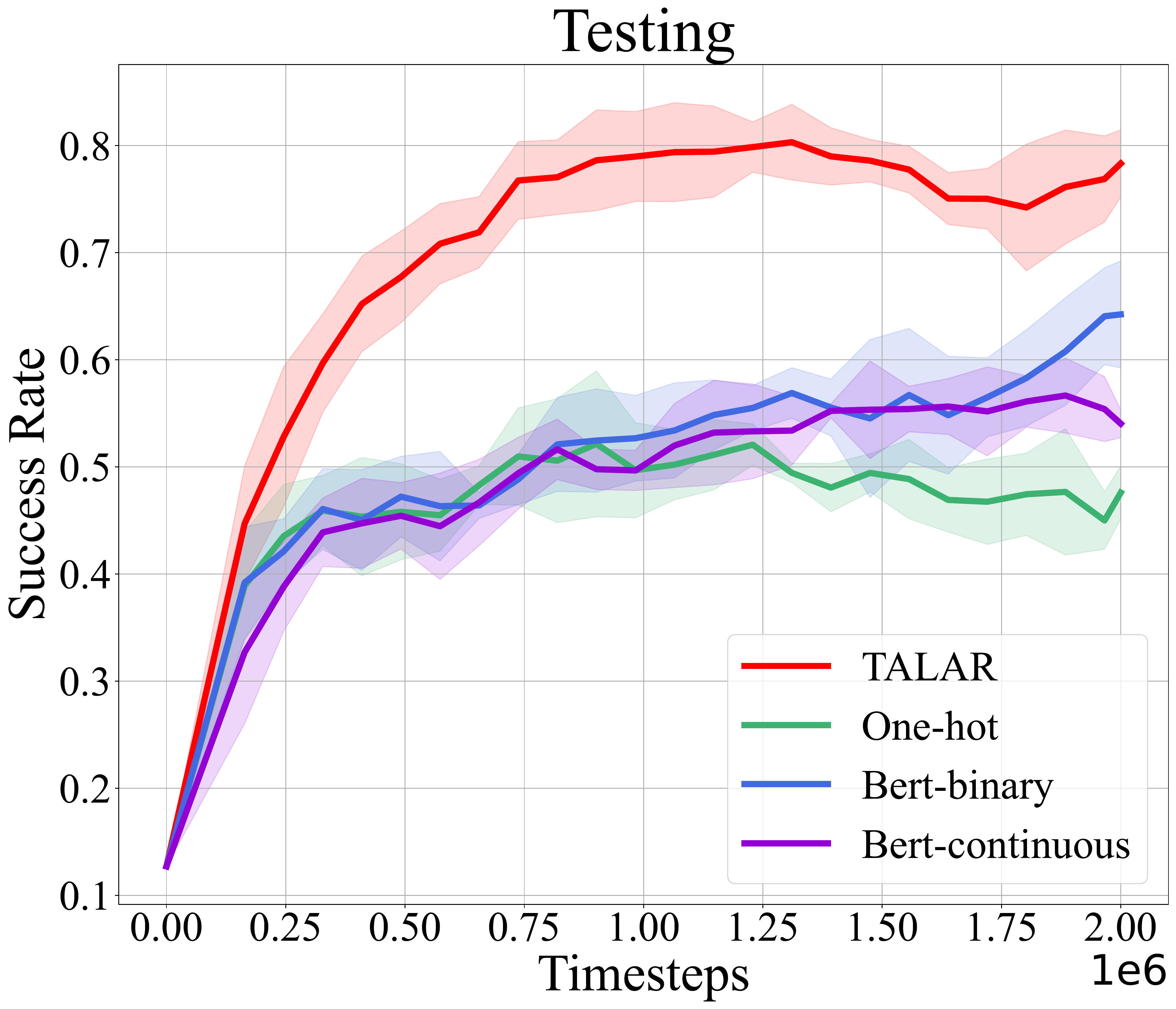}
}
\subcaptionbox*{}{
\includegraphics[width=0.315\textwidth]{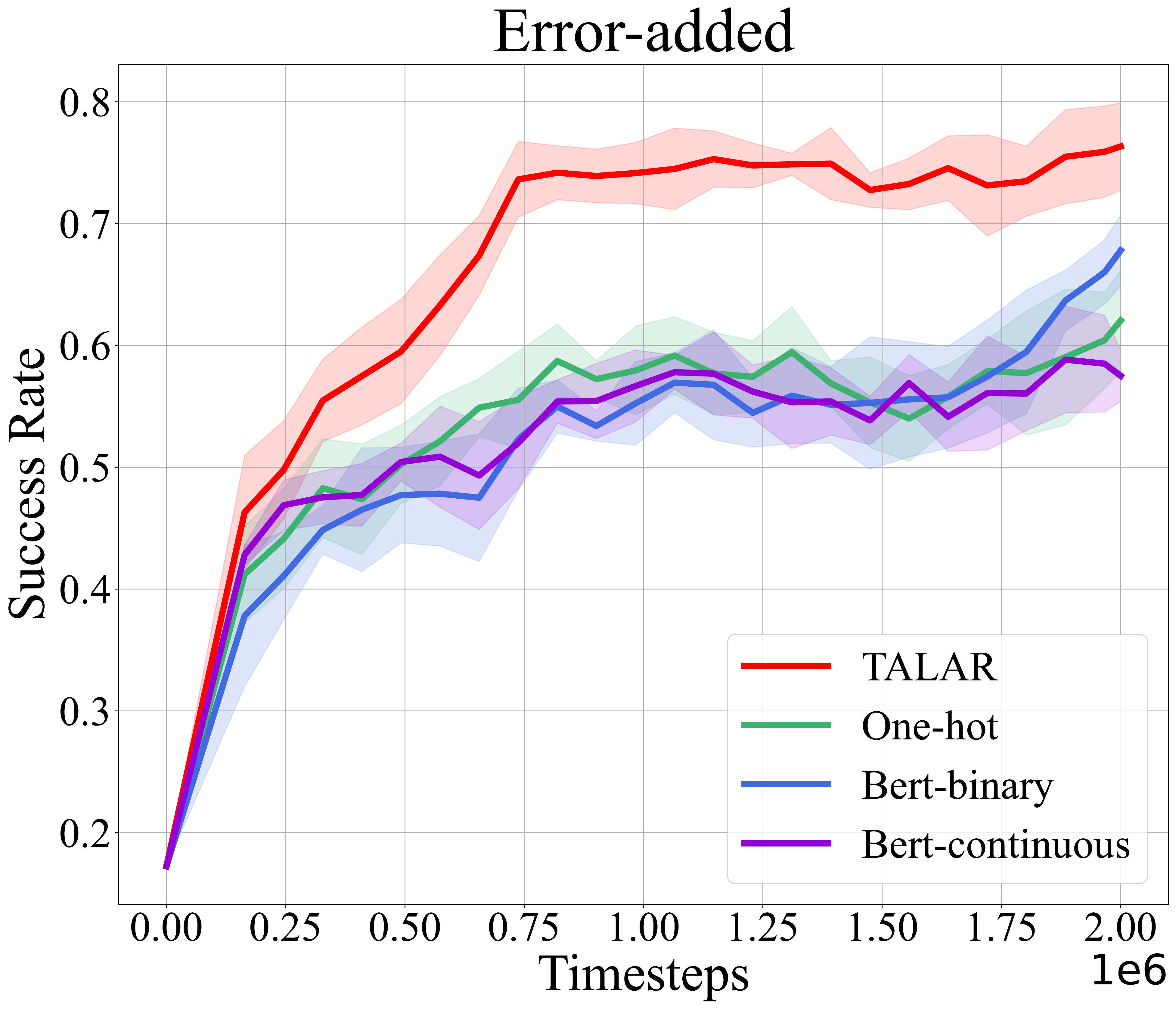}
}
\caption{Training curves of different methods on three NL instruction datasets. The x-axis represents the total timesteps agent interacts with the environment, and the y-axis represents the success rate of completing instructions. The shaded area stands for the standard deviation over five random trials.}
\label{fig:ifp_results}
\end{figure}

\subsection{Performance of Instruction-Following Policy}
\label{sec:exp_ifp}
.5With the optimized translator, we train an instruction-following policy in the CLEVR-Robot task, following the training process elaborated in Section \ref{sec:method_policy_training}. We first introduce three datasets of different NL expressions. (1) \textbf{Training set} contains nine NL sentence patterns (i.e., 720 NL instructions) for policy learning. All agents only interact with the training dataset when optimizing the policy. (2) \textbf{Testing set} contains nine NL sentence patterns (i.e., 720 NL instructions) that are different from the training set. (3) \textbf{Error-added set} contains the exact 720 NL instructions as the training set, with the addition of errors to each NL instruction, such as the word [the] being omitted. See Appendix \ref{appendix:implmentation_tasks} for information regarding these three datasets and evaluation tasks.

\textbf{Baselines for comparison}. We consider multiple baselines that are built upon OIL architecture (i.e., standard NLC-RL): 
(1) \textbf{Bert-binary} processes the NL with a pre-trained Bert LM. The language encoding from the Bert is processed to a binary vector by a fully-connected network. This binary vector's size equals TL generated by \methodname. 
To ensure the differentiability, we use a reparameterization trick \cite{reparameterization} that converts continuous vector to binary vector.
(2) \textbf{Bert-continuous} is similar to Bert-binary, except that it replaces the binary vector with a continuous vector of the same size.
(3) \textbf{One-hot} encodes the representation of all possible NL instructions (including training, testing and error-added) to a three-dimensional tensor, where each instruction has its position.

\begin{table}[t]
    \centering
    \caption{A summary of the final success rate (\%) in instruction-following task with different sets of NL expressions. The results are averaged over $5$ seeds, and each data is evaluated for 40 episodes.}
    \begin{tabular}{c|c|c|c}
    \toprule
    \diagbox{Method}{Dataset}  & Training & Testing & Error-added  \\   
    \midrule
    \methodname & \textbf{99.9}$\pm$ \textbf{0.1} & \textbf{78.3} $\pm$ \textbf{3.1} & \textbf{76.3} $\pm$ \textbf{3.6} \\ \midrule
    One-hot & 86.5 $\pm$ 3.0 & 47.6 $\pm$ 2.5 & 62.1 $\pm$ 4.1 \\ \midrule
    Bert-binary & 64.0 $\pm$ 2.5 & 64.2 $\pm$ 5.0 & 67.8 $\pm$ 2.9 \\ \midrule
    Bert-continuous & 60.7 $\pm$ 1.9 & 54.0 $\pm$ 1.3 & 57.5 $\pm$ 2.1 \\ \midrule
    \end{tabular}
    \label{tab:overall_sr}
\end{table}

\textbf{Experimental results.} Figure \ref{fig:ifp_results} presents the training curves on the instruction-following task with various NL instruction datasets, and Table \ref{tab:overall_sr} summarizes the final success rates of all methods. Overall, \methodname~acquires a better instruction-following policy that increases the success rate by 13.4\% relative to OIL baselines and adapts to previously unseen expressions of NL instruction. On the training NL instruction set, \methodname~achieves a success rate greater than 99\% within 2M timesteps, significantly faster than the two Bert-based baselines.
Combined with the results in Section \ref{sec:exp_language_generation}, the translator can effectively convert diverse NL expressions to a unique TL representation, which enables efficient policy learning.
Besides, \methodname~achieves a success rate greater than 76\% in both testing and error-added sets, demonstrating greater capacity than baselines. 
While One-hot performs adequately on the training NL set, its generalization to the testing and error-added NL sets is limited. 
Two Bert-based baselines improve more slowly than \methodname~on the training NL instruction set, which can be attributed to the fact that they simultaneously train a policy while acquiring skills and understanding NL. 
Bert's encoding of different NL expressions can be highly diverse, which adds complexity to OIL baselines to solve RL tasks; consequently, Bert-based baselines improve more slowly than \methodname~during the training process.

\subsection{TL as an Abstraction for Hierarchical RL}
\label{sec:exp_hrl}

\begin{figure}[t]
\centering
\subcaptionbox{An example of successful object arrangement.\label{fig:env_arrangement}}{
\includegraphics[width=0.35\textwidth]{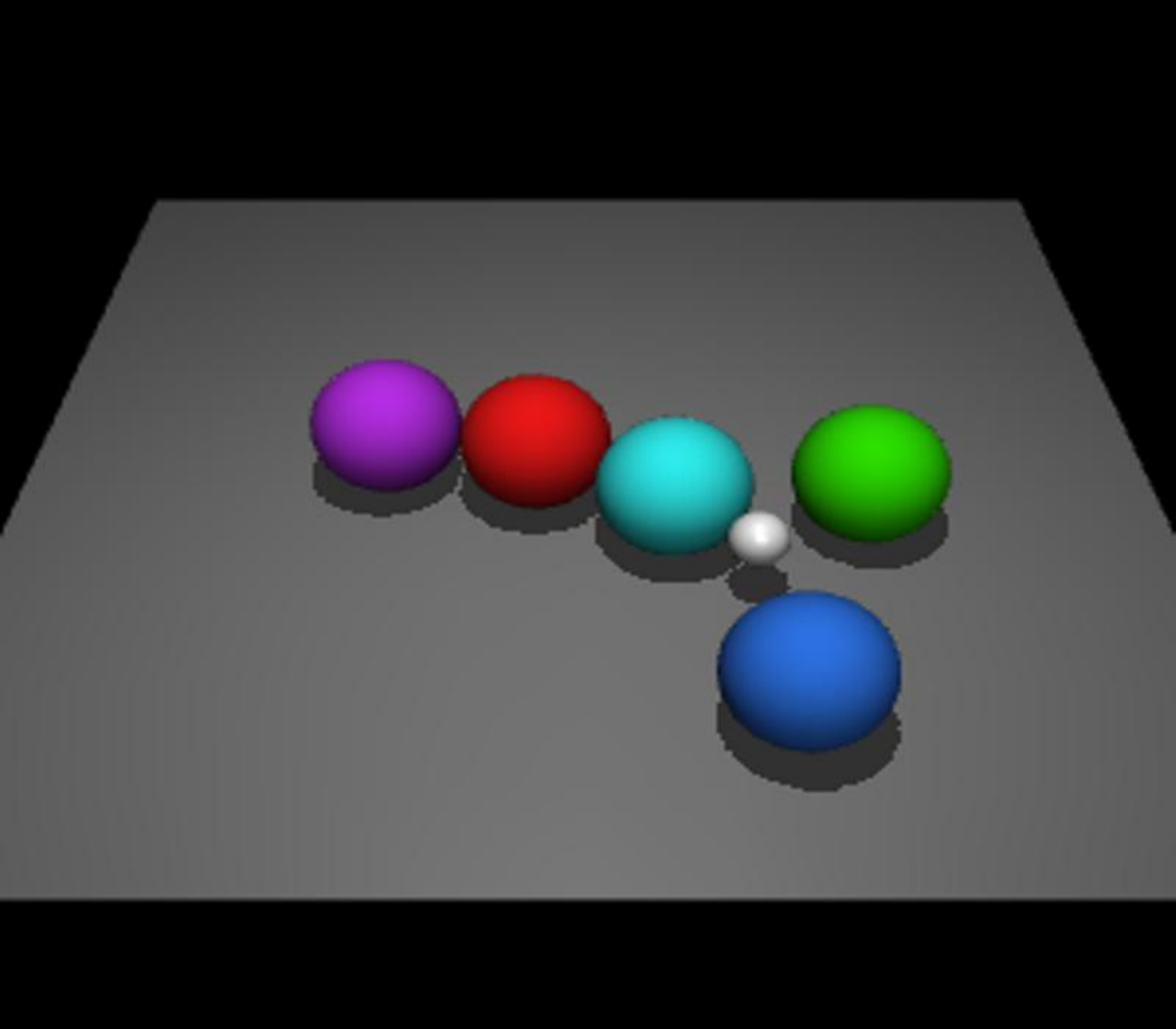}

}
\subcaptionbox{Success rate curves during the training process.\label{fig:hrl_training}}{
\includegraphics[width=0.35\textwidth]{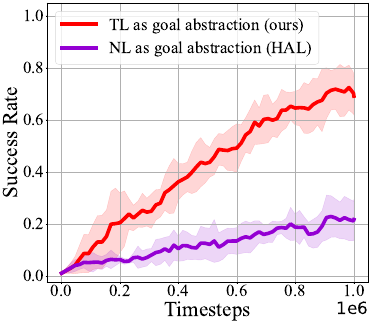}
}
\caption{Experiments in a long-term object arrangement task. (a) A snapshot of successful object arrangement, which aims to arrange objects to satisfy all $10$ human instructions at the same time. (b) Training curves of different goal abstractions in object arrangement.} 
\end{figure}

Previous experiments demonstrate that the resulting TL is a unique representation of the various NL expressions, which assists a policy in efficiently learning to follow NL instructions.
In this section, we further explore the applicability of generated TL by examining if it can serve as an effective goal abstraction for hierarchical RL. Specifically, we train a high-level policy outputting a TL, instructing the IFP to complete a low-level task. 
We consider a baseline \textbf{HAL} \cite{language_as_abstraction} for comparison. HAL takes advantage of the compositional structure of NL and directly makes decisions on the NL level. Following its original implementation, the high-level policy of HAL outputs the index of NL instruction. To ensure a fair comparison, HAL uses the IFP trained by \methodname~as a low-level policy in our experiment. The high-level policies are trained with the PPO algorithm.
We consider a long-term task based on the CLEVR-Robot environment, namely object arrangement, as shown in Figure \ref{fig:env_arrangement}. The task objective of object arrangement is to arrange objects to satisfy all ten constraints that are implicitly contained in the task. Figure \ref{fig:hrl_training} presents the comparison results. The high-level policy that uses TL as a low-level goal abstraction performs significantly better than that using NL in terms of improving speed. This result shows that TL can be a helpful goal abstraction naturally compatible with hierarchical RL.

\subsection{Ablation Study}
\label{sec:exp_ablation}

We further conduct ablation experiments to verify how each component affects the performance of \methodname.

\begin{figure}[htbp]
  \centering
  \subcaptionbox{TL-predicate\label{fig:tl_predicate_rep}}{\includegraphics[width=0.3\linewidth]{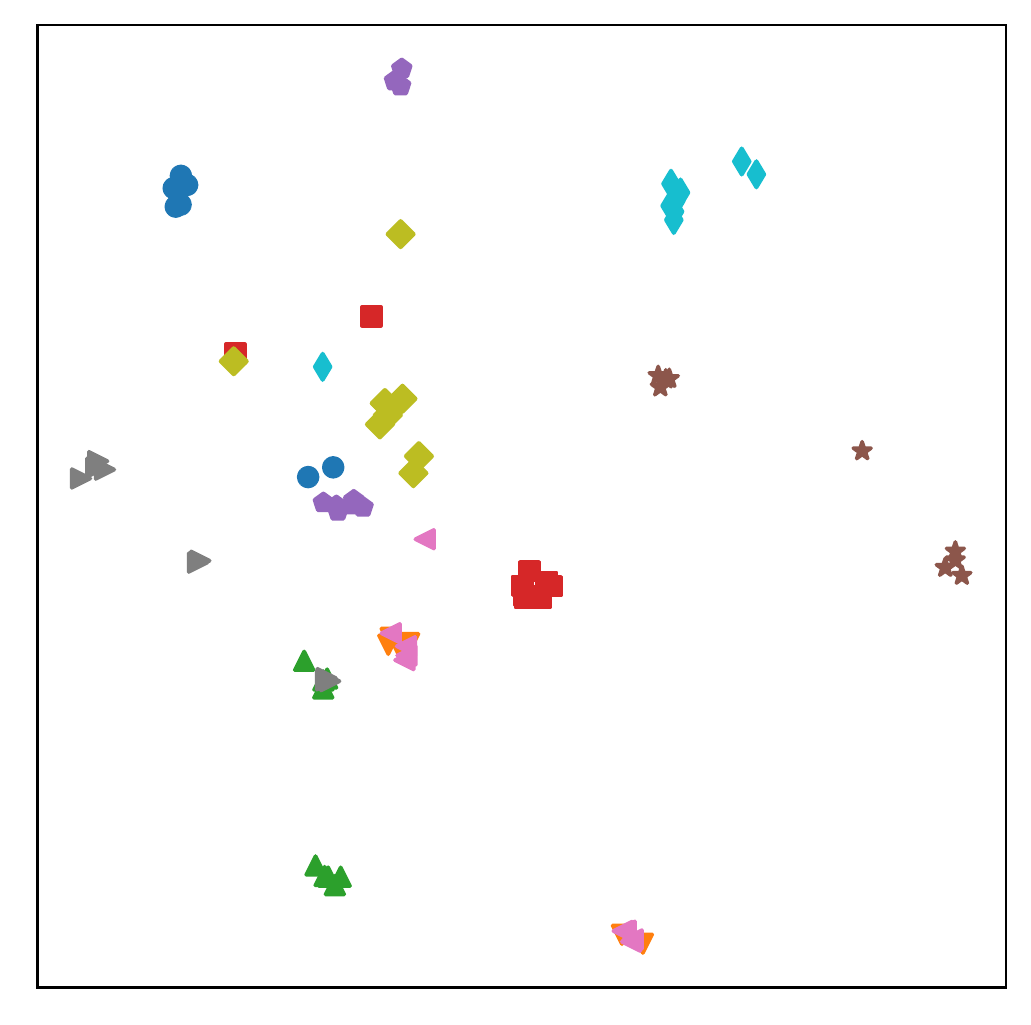}}
  \subcaptionbox{TL-binary\label{fig:tl_binary_rep}}{\includegraphics[width=0.3\linewidth]{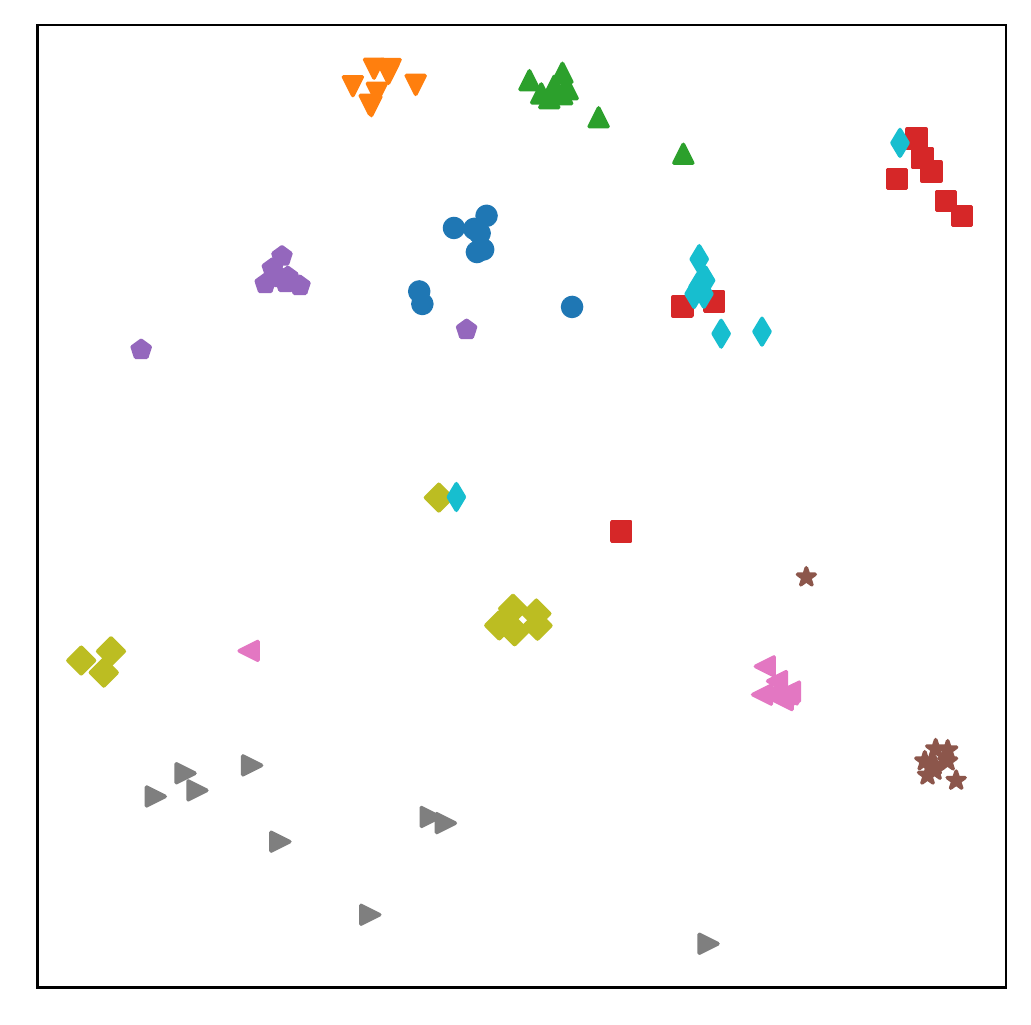}}
  \subcaptionbox{TL-continuous\label{fig:tl_cont_rep}}{\includegraphics[width=0.3\linewidth]{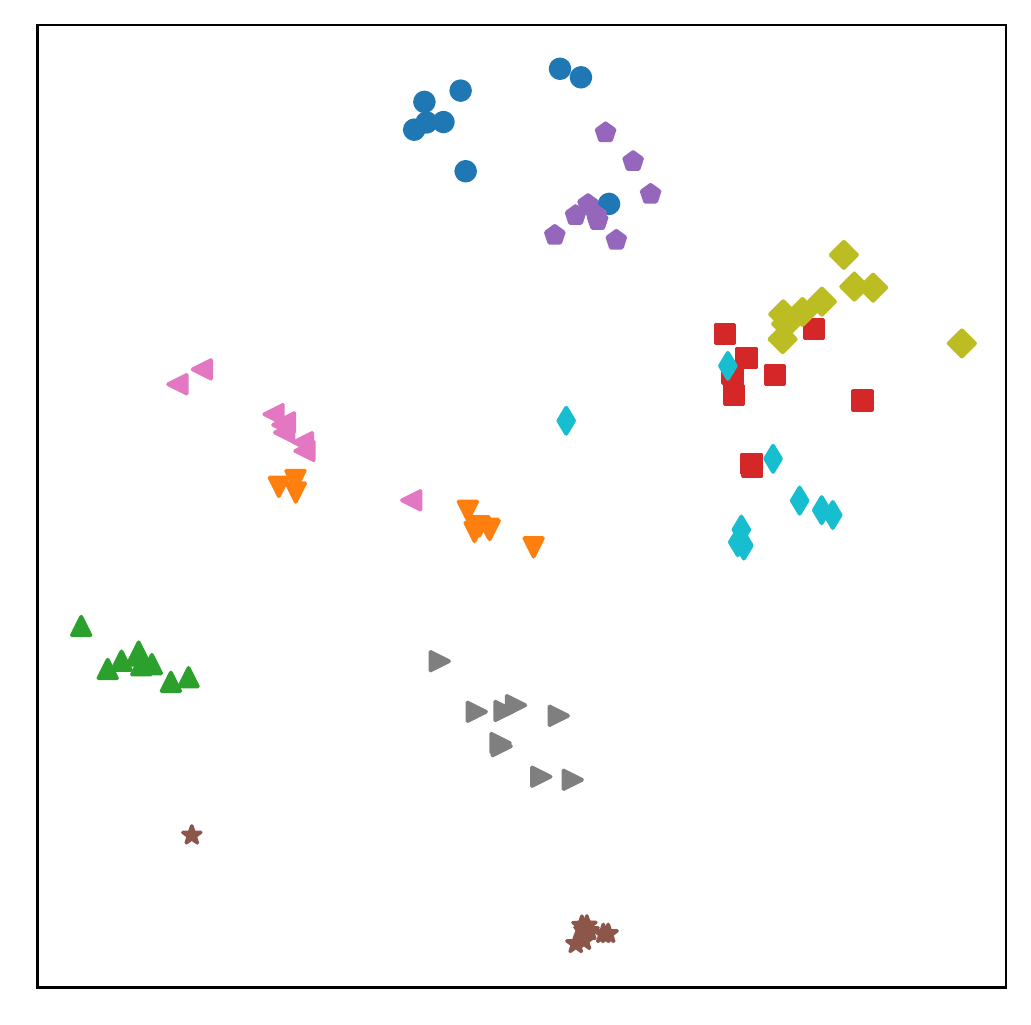}}
  \caption{The t-SNE projections of the task language in different kinds of representations, with the points generated from the testing NL instruction set.}
  \label{fig:tsne_training_tl_train_set}
\end{figure}

\textbf{Predicate representation.} 
To evaluate the efficacy of predicate representation in TALAR, we replace it with binary/continuous vectors and derive two representations of TL, TL-binary and TL-continuous. 
In TL-binary/TL-continuous, a multi-layer perception network outputs the TL vector directly. 
Figure \ref{fig:tsne_training_tl_train_set} shows the t-SNE projection of different kinds of TL representations. The results are computed based on the testing NL instruction set. For each human instruction with different NL expressions, the points produced by TL-predicate are more concentrated than those of TL-binary and TL-continuous. These results demonstrate the effectiveness of predicate representation for developing TL. 
Besides, Table \ref{tab:ablation_tl_rep} displays the final success rate of IFPs trained with various TL representations. Overall, all three types of representations are adequate for achieving a high task success rate on the training NL instruction set. However, predicate representation is more adaptable to unseen NL expressions and has a higher task success rate on testing/error-added NL instruction sets.

\begin{table}[htbp]
    \centering
    \caption{Comparisons of different representations of TL. The success rate (\%) is averaged over $5$ seeds.
    }
    \begin{tabular}{c|c|c|c}
    \toprule
    \diagbox{TL rep.}{Dataset} & Training & Testing & Error-added \\   
    \midrule
    Predicate & \textbf{99.9} & \textbf{78.3} & \textbf{76.3} \\ \midrule
    Binary & \textbf{99.8} & 77.1 & 75.7 \\ \midrule
    Continuous  & \textbf{99.8} & \textbf{78.1} & 66.3 \\ \midrule
    \end{tabular}
    \label{tab:ablation_tl_rep}
\end{table}

\begin{figure}[htbp]
\centering
\subcaptionbox*{}{
\includegraphics[width=0.35\textwidth]{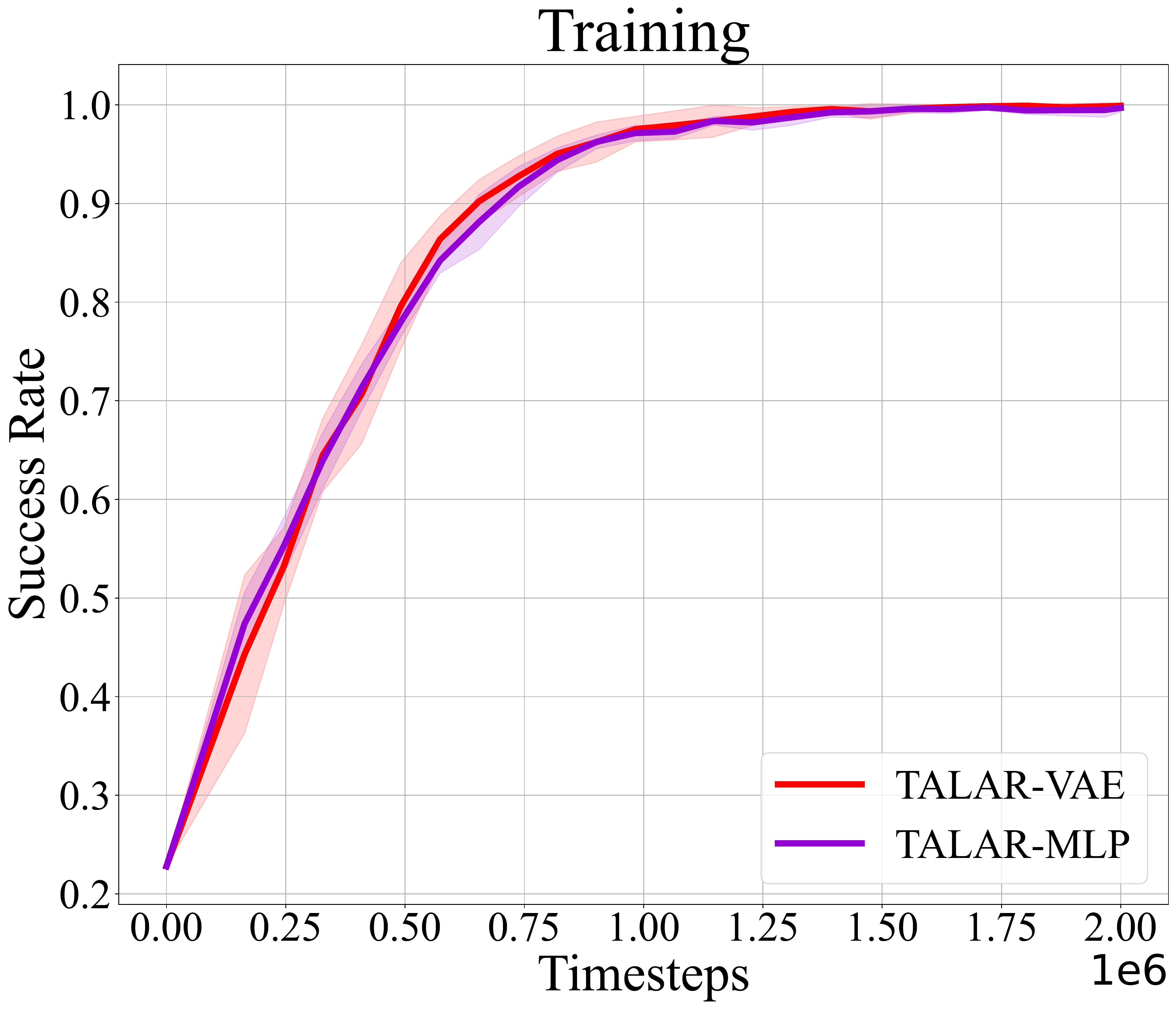}
}
\subcaptionbox*{}{
\includegraphics[width=0.35\textwidth]{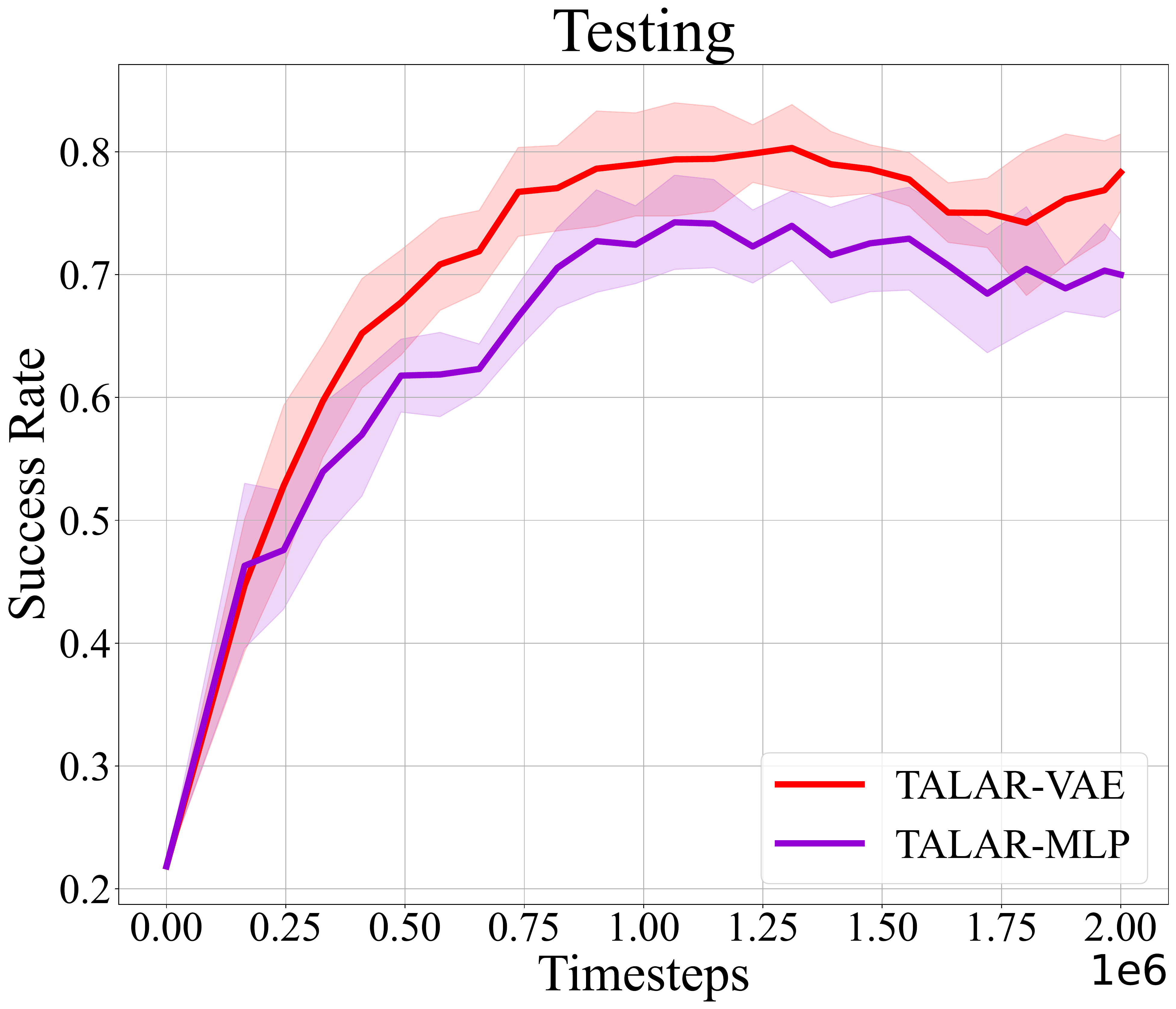}
}
\caption{Training curves of IFP with different structures of the translator. See Appendix \ref{appendix:extra_mlp} for the training curves on the error-added set.} 
\label{fig:mlp_translator}
\end{figure}

\textbf{VAE of the translator.}
\methodname~employs a VAE for translator training. To demonstrate its efficacy, we introduce \methodname-MLP, which replaces the VAE in \methodname~with a multi-layer perception (MLP) network and trains the translator using supervised learning loss. 
Then, we train an IFP using the MLP translator, and the experiment results are depicted in Figure \ref{fig:mlp_translator}. Despite having comparable success rates on the training set, \methodname-VAE outperforms \methodname-MLP on the testing NL instruction set by 8.4\% of success rate. This result suggests that VAE is advantageous when training a translator to generalize unseen NL expressions.

\begin{figure}[htbp]
  \centering
  \subcaptionbox{Training\label{fig:heatmap_train}}{
  \includegraphics[width=0.25\textwidth]{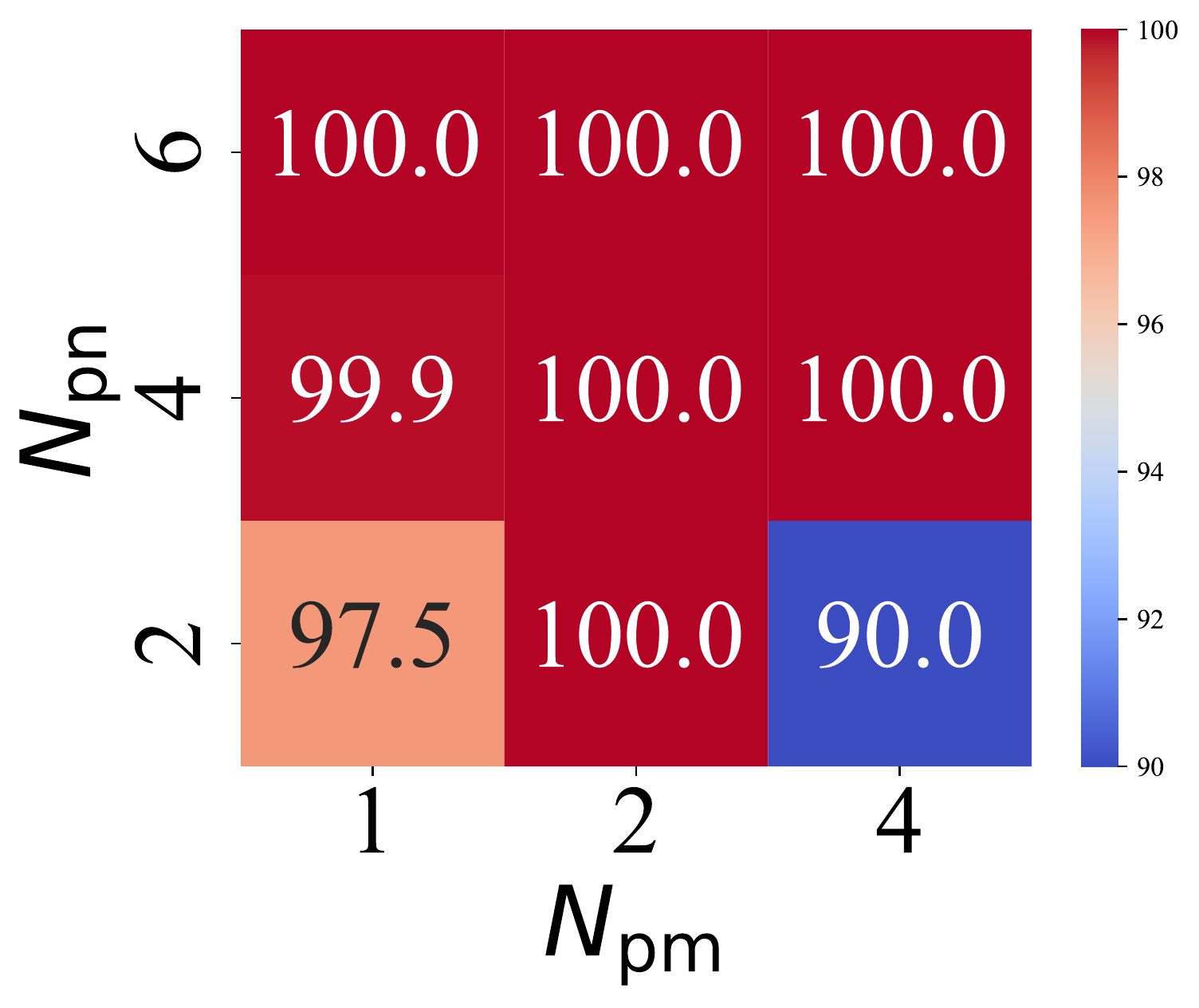}
  }
  \subcaptionbox{Testing}{
  \includegraphics[width=0.25\textwidth]{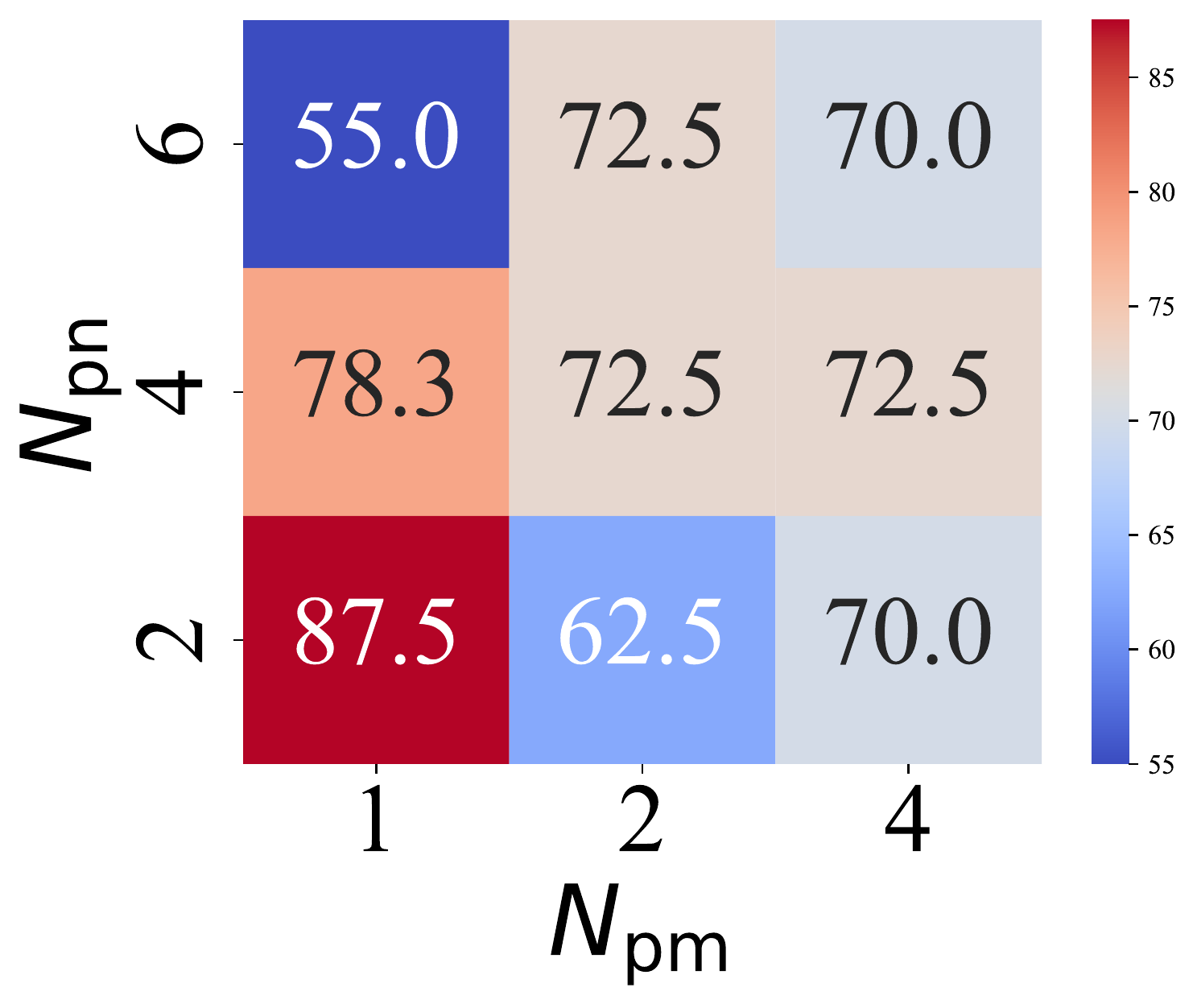}
  }
  \subcaptionbox{Error-added\label{fig:heatmap_error}}{
  \includegraphics[width=0.25\textwidth]{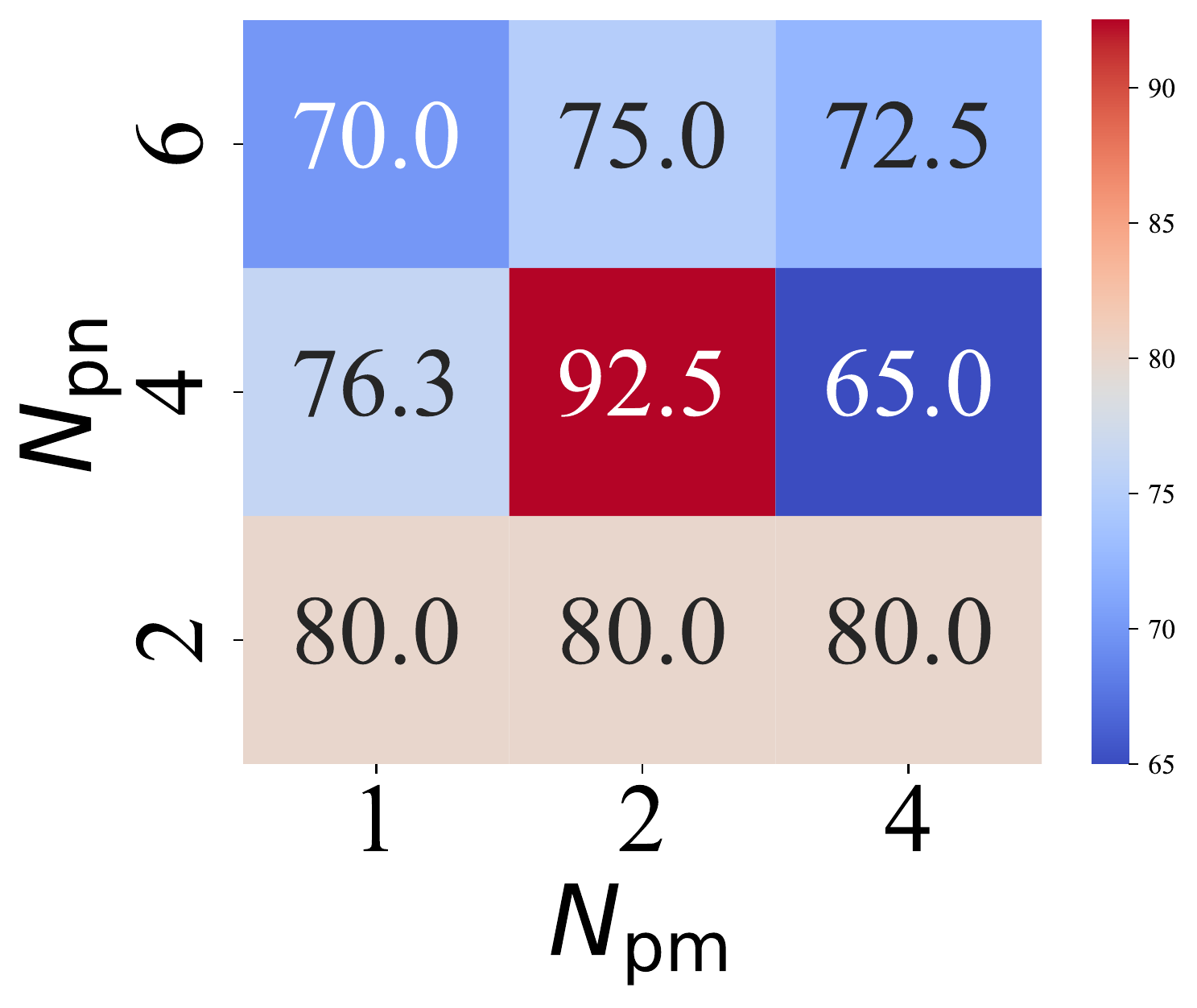}
  }
  \caption{Ablation study on the number of predicate modules/networks. The values in the heat map represent the success rate of IFPs trained for 2M timesteps, with different parameter configurations of $N_{\rm pm}$ and $N_{\rm pn}$. 
  } 
  \label{fig:ablation_number}
\end{figure}

\textbf{Number of predicate modules and predicate networks.}
We conduct experiments to examine how the number of predicate modules/networks (i.e., $N_{\text{pm}}$ and $N_{\text{pn}}$) in \methodname~affects the performance. In our experiments, $N_{\text{pm}}$ is selected from [1, 2, 4], while $N_{\text{pn}}$ is selected from [2, 4, 6]. Figure \ref{fig:ablation_number} shows the experiment results. In general, greater $N_{\text{pm}}$ and $N_{\text{pn}}$ result in improved performance on the training NL instruction set (see Figure \ref{fig:heatmap_train}). However, the experimental performance on the testing/error-added set is quite the opposite, as shown in Figure \ref{fig:heatmap_error} when $N_{\text{pm}}=4$ and $N_{\text{pn}}=6$. 
This result could be attributed to the fact that, as the number of predicate modules and networks increases, the representation of TL becomes more complex (i.e., the vector size increases), making it more difficult for the policy to follow the TL. 
Besides, we also observe that, within a specific range of values ($N_{\text{pm}} \leq 2$ and $N_{\text{pn}} \leq 4$), larger $N_{\text{pn}}$ and $N_{\text{pm}}$ would bring a better performance. These results serve as a guide for selecting appropriate hyper-parameters.
Due to space constraints, experiments on \methodname~involving a variety of argument networks are presented in Appendix \ref{appendix:diff_num_of_arg_net}.

\section{Conclusion}
This paper focuses on the topic of NLC-RL.
We suggest that NL is an unbounded representation of human instruction, thereby imposing a substantial additional burden on the policy when solving RL tasks. 
To alleviate the burden, we investigate a new IOL scheme for NLC-RL by developing TL, which is task-related, and a unique representation of human instruction.
Through our experiments, we verify that the resulting TL can uniquely represent human instructions with diverse NL expressions and is interpretable to some extent. Besides, the policy following TL can quickly learn to complete the instructions with a high success rate and adapts to previously unseen NL expressions. Moreover, the resulting TL is an effective goal abstraction of a low-level policy that serves as the basis for hierarchical RL.

Although \methodname~can effectively train a competent instruction-following policy, there are limitations. 
\methodname~develops the task language using a static task dataset and, therefore, can not be directly applied to an open environment task. It is possible to mitigate this issue by dynamically extending the task dataset and fine-tuning the TL generator/translator during the policy learning process.
Besides, \methodname~requires a manual reward function for policy training, which may be inaccessible if the reward design is complex. Fortunately, there have been well-validated methods for solving sparse reward problems \cite{HER,overcome_sparse_reward,overcome_sparse_reward_2}, which is an effective substitute for the manual reward function. 
Finally, it would be interesting to involve the basic properties of predicate relationships (such as transitivity, reflexivity, and symmetry) when training the TL generator, which makes the resulting TL more meaningful and self-contained. 
We hope future research will investigate these intriguing questions and make strides toward training agents that interact with humans more effectively.

\clearpage

\bibliographystyle{abbrvnat}
\bibliography{references}

\newpage
\appendix
\label{Appendix}
\onecolumn

\begin{center}
    \Large{\textbf{APPENDIX}}
\end{center}
    
\section{Discussion}
\label{appendix:discussion}

This section will present examples to illustrate the predicate representation better.

\subsection{What Is the Predicate Representation?} 
Overall, predicate representation utilizes the form of predicate expressions, which can model various relationships in the tasks. Specifically, in this paper, we utilize a discrete binary vector to represent the truth value of multiple anonymous predicates and the arguments of these predicates.
For example, in a predicate representation vector, [1, \textcolor{red}{0, 1, 0}, \textcolor{blue}{1, 0, 0}], the first code [1] stands for the value of an anonymous predicate is True, the red and blue codes stand for the indexes of two arguments of this predicate, respectively.

\subsection{What Are the Advantages of the Predicate Representation?}
In the main body (Section \ref{sec:method}), we have mentioned that predicate representation is expressive.
We suggest that such expressiveness comes from two key advantages of the predicate representation: \textit{compositional structure} and \textit{interpretability}.
We first discuss the \textbf{compositionality}, which refers to the ability of a language to denote novel composite meanings. For example, if a language can represent \textit{blue circle} and \textit{red square}, it can represent \textit{blue square} as well. This language has a compositional structure.
The compositionality is seen both as a fundamental feature of natural language and as a pre-condition for a language to generalize at scale 
\cite{compos_structure}. Predicate representation naturally has 
compositional structure due to the fact that it can denote the 
\begin{wrapfigure}[16]{r}{0.35 \textwidth}
\vspace{-1em}
\includegraphics[width=0.35 \textwidth]{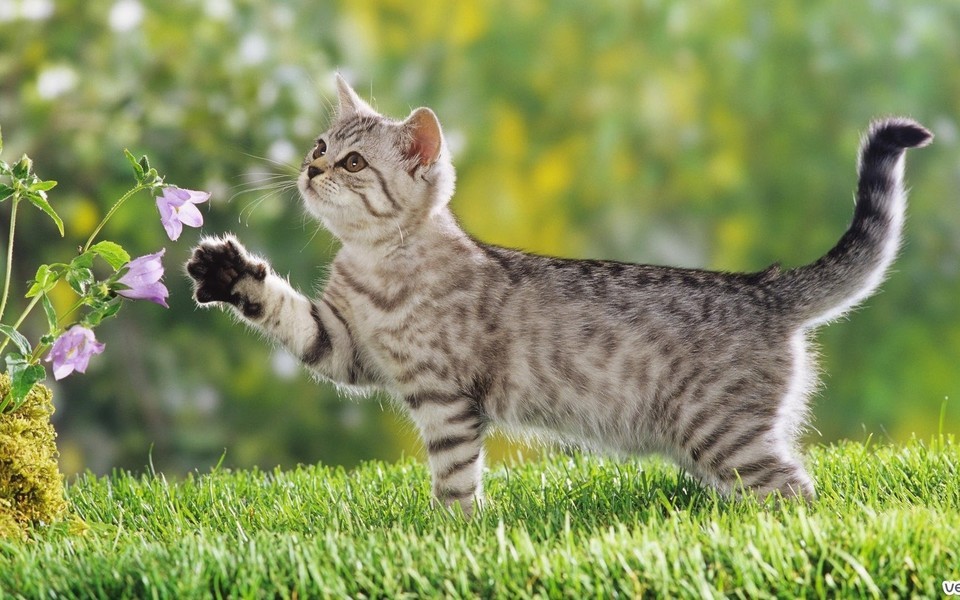}
\caption{An illustration of the interpretability of predicate representation. For an anonymous predicate expression: \texttt{Pred(cat,grassland)=True}, we can guess that \texttt{Pred} represents [above].}
\label{fig:cat}
\end{wrapfigure} 
composite meanings by changing the truth value of predicates.

Next, we talk about the \textbf{interpretability}.
Predicate representation uses multiple predicate expressions, such as \texttt{Pred(a1,a2)}, to describe the relationships in a specific environment. Previous works have found that, even if the predicate symbols are machine-generated by a learning system and anonymous to humans, they still offer \textit{some extent of interpretability}. For example, if a learning system generates a predicate expression \texttt{Pred(cat,grassland)=True} for Figure \ref{fig:cat}, we can guess that \texttt{Pred} means [stand] or [above]. When more figures and predicate expressions are provided, the actual meaning of these anonymous predicate symbols would be more apparent to humans.

In conclusion, we implement the IOL architecture utilizing predicate representation because nearly all tasks contain relationships. For instance, an Atari game \cite{atari} contains a relationship between the agent and the antagonist, a first-person shooter game contains a relationship between the gun and the bullets, a stock trading task contains a relationship between stocks and price, etc. 
Even if some tasks are extremely abstract and may have no relationships, the predicate representation can be replaced with TL in binary or continuous representations, which have demonstrated their ability to develop a task language in our experiments (Section \ref{sec:exp_ablation}).

\section{Additional Related Work}
\label{appendix:add_related_work}

\textbf{Hierarchical Reinforcement Learning} (HRL) approaches are promising tools for solving complex decision-making tasks \cite{hrl_survey}. 
HRL solves the problem by decomposing it into simpler sub-tasks using a hierarchy of policies learned by RL \cite{vezhnevets2017feudal, jong2008utility, nachum2019does}. For example, to put a drink into a fridge, you will (1) take up the drink, (2) open the fridge door, (3) put down the drink, and (4) close the fridge door. Putting a drink into a fridge is a long-term task, while these four are sub-tasks.
Typically, a high-level policy is trained to decompose the main task into sub-tasks (three steps in the example), and low-level policies (or single policy) are trained to complete these sub-tasks. 
A \emph{core problem} in HRL is how to represent the goal of the sub-task (i.e., goal representation) that the low-level policy follows. Previous works have investigated both concrete and abstract goal representations of a sub-task. Concrete goal representation can be the \textit{target position} to reach \cite{hrl_target_pos}, or a \textit{target state} \cite{hrl_target_state}. As for abstract goal representation, it can directly be natural language \cite{language_as_abstraction} or an encoding of action primitives \cite{skill_prior}. 
Back to the topic of this paper, language is a natural goal representation for the sub-tasks, such as \textit{open the fridge door} in the above example. 
Through our experiments (Section \ref{sec:exp_hrl}), we observe that the resulting TL is an effective abstraction of the goal of the sub-task, which demonstrates the applicability of TL to work together with HRL.

\section{Algorithm Descriptions}
\label{appendix:algorithm}
Algorithm \ref{alg:tl_generator}, \ref{alg:translator}, and \ref{alg:policy_training} present the training procedures of the TL generator, the translator and the instruction-following policy, respectively.

\begin{algorithm}[htbp]
    \caption{Training procedure of the TL generator.}
    \textbf{Input}: task dataset $\gD = \{(s,s',\nl)_i\}$, pre-trained Bert model $b$. \\
    \textbf{Output}: the optimized TL generator.
    \begin{algorithmic}[1] 
        \STATE Initialize the TL generator $g_\theta$.
        \WHILE{training not complete}
        \STATE Sample a batch of data from $\gD$.
        \STATE Update $\theta$ to minimize the MLM loss (Eq.(\ref{eq:mlm_loss})).
        \ENDWHILE
        \STATE \textbf{return} the optimized TL generator $g_\theta$.
    \end{algorithmic}
    \label{alg:tl_generator}
\end{algorithm}

\begin{algorithm}[htbp]
    \caption{Training procedure of the translator.}
    \textbf{Input}: task dataset $\gD = \{(s,s',\nl)_i\}$, the optimized TL generator $g_\theta$, and the pre-trained Bert model.  \\
    \textbf{Output}: the optimized TL generator.
    \begin{algorithmic}[1] %
        \STATE Initialize the translator $t_{\phi_1,\phi_2}$ with parameters $\phi_1$ and $\phi_2$.
        \WHILE{training not complete}
        \STATE Sample a batch of data $\{(s,s',\nl)_j\}$ from $\gD$.
        \STATE // Compute the target task language which VAE aims to recover.
        \STATE Calculate the task language $\tl=g_\theta(s,s')$. 
        \STATE Update $\phi_1,\phi_2$ to minimize the VAE loss (Eq.(\ref{vae_loss})).
        \ENDWHILE
        \STATE \textbf{return} the optimized translator $t_{\phi_1,\phi_2}$.
    \end{algorithmic}
    \label{alg:translator}
\end{algorithm}

\begin{algorithm}[htbp]
    \caption{Training procedure of the instruction-following policy.}
    \textbf{Input}: the optimized translator $t_{\phi_1,\phi_2}$.  \\
    \textbf{Output}: the optimized instruction-following policy.
    \begin{algorithmic}[1] %
        \STATE Initialize the policy function $\pi$, and the value function.
        \WHILE{training not complete}
        \STATE Sample a NL instruction $\nl$ from the environment.
        \STATE Generate corresponding task language $\widetilde{\tl}=t_{\phi_1,\phi_2}(\nl)$.
        \STATE // Collecting samples
        \WHILE{episode not terminal}
        \STATE Observe current state $s_t$.
        \STATE Execute action $a_t \sim \pi(\cdot|s_t,\widetilde{\tl})$, and receive a reward $r_t$ from the environment.
        \ENDWHILE
        \STATE // Training
        \STATE Update the policy and value functions based on the samples collected from the environment.
        \ENDWHILE
        \STATE \textbf{return} the optimized policy $\pi$.
    \end{algorithmic}
    \label{alg:policy_training}
\end{algorithm}

\section{Implementation Details}
\label{appendix:implmentation}

In our experiments, we utilize the open-sourced RL repository, stable-baselines3 \cite{stable-baselines3}, to implement the RL training. All experiments are run for five times with different random seeds. We will next introduce the tasks for evaluation and the hyper-parameters used in our experiments.

\subsection{Tasks for Evaluation}
\label{appendix:implmentation_tasks}

\textbf{Instruction following task.}
As depicted in Figure \ref{fig:env}, our experiments are conducted in a CLEVR-Robot environment where an agent manipulates five balls to complete human instructions. The observation space is $s\in \mathbb{R}^{10}$, which represents the location of each object, and $|A| = 40$ corresponds to selecting and pushing an object in one of the eight cardinal directions. At each timestep, the reward is equal to (distance to the target position at the previous step) minus (distance to the target position at the current step), with +5 for success and -5 for failure. Some previous OIL-based methods attempt to train the IFP in a sparse reward environment \cite{ins_fol_sparse_reward_1,ins_fol_sparse_reward_2}. We relax the sparse reward constraint to dense reward for two reasons: (1) We are more concerned with exploring a new learning scheme for NLC-RL than with resolving the sparse reward problem; (2) Most NLC-RL approaches address the sparse reward problem using HER \cite{HER}, which relabels the origin goal in the trajectory with the actual achieved state. Nonetheless, the relabeling procedure is contingent on the transformation from the state to the goal, which necessitates additional human annotation or program design.

We use eighteen different NL sentence patterns to express each human instruction. Take a human instruction, blue ball to the right of the green ball, for example, its corresponding NL instructions (i.e., or eighteen NL sentence patterns) can be one of the:
\begin{itemize}[itemsep=2pt,topsep=0pt,parsep=0pt]
    \item ---(\textbf{Training set})--- 
    \item Push the blue ball to the right of the green ball.
    \item Can you push the red ball to the right of the green ball?
    \item Can you help me push the red ball to the right of the green ball?
    \item Is the red ball right of the green ball?
    \item Is there any red ball right the green ball?
    \item The red ball moves to the right of the green ball.
    \item The red ball is being pushed to the right of the green ball.
    \item The red ball is pushed to the right of the green ball.
    \item The red ball was moved to the right of the green ball.
    \item ---(\textbf{Testing set})--- 
    \item Keep the red ball right of the green ball.
    \item Move the red ball to the right of the green ball.
    \item Can you move the red ball to the right of the green ball?
    \item Can you keep the red ball to the right of the green ball?
    \item Can you help me move the red ball to the right of the green ball?
    \item Can you help me keep the red ball to the right of the green ball?
    \item The red ball is being moved to the right of the green ball.
    \item The red ball is moved to the right of the green ball.
    \item The red ball was pushed to the right of the green ball.
\end{itemize}

We also consider a set of errors in the NL sentence:
\begin{itemize}[itemsep=2pt,topsep=0pt,parsep=0pt]
    \item Missing a [the].
    \item Incorrect use of prepositions, e.g., using [\textit{on front of}] to replace [\textit{in front of}].
    \item Incorrect use of phrase, including \textit{in behind of}, \textit{in left of}, and \textit{in right of}.
    \item Oversimplifying the expression, e.g., move red ball right green ball.
\end{itemize}

In our experiments, there are three kinds of datasets: training, testing, and error-added. 
In the training set, human instructions are expressed using nine NL sentence patterns, while the remaining nine NL sentence patterns are used in the testing set. The error-added set utilizes the same NL nine sentence patterns as the training set, but each sentence has at least one of the errors listed above.
At the start of each trajectory, the environment randomly samples a human instruction and an NL sentence pattern to express the human instruction. 
By constructing these three datasets, we simulate the scenario in an open environment where different individuals instruct the robots using their linguistic preferences.

\textbf{High-level object arrangement task.}
Object arrangement task is proposed by \cite{language_as_abstraction}, which aims to rearrange the objects in the environment to satisfy all ten implicit constraints. 
At the start of a trajectory, the environment resets the position of all balls to a random location. At each time step, the reward agent receives a reward equal to (number of constraints satisfied at current timestep) - (number of constraints satisfied at previous timestep). Following \cite{language_as_abstraction}, the precise arrangement constraints are: (1) red ball to the right of purple ball; (2) green ball to the right of red ball; (3) green ball to the right of cyan ball; (4) purple ball to the left of cyan ball; (5) cyan ball to the right of purple ball; (6) red ball in front of blue ball; (7) red ball to the left of green ball; (8) green ball in front of blue ball; (9) purple ball to the left of cyan ball; (10) blue ball behind the red ball.

\subsection{Hyper-Parameters}
The hyper-parameters for implementing \methodname~are presented in Table \ref{tab:hyper-parameters}.

\begin{table}[htbp]
    \centering
    \caption{Hyper-parameters in our experiments.}
    \begin{tabular}{l|l}
    \toprule
    \textbf{Hyper-parameters} & \textbf{Value}  \\   \toprule
    $N_{\text{a}}$ & 2 \\ \midrule
    $N_{\text{pn}}$ & 4 \\ \midrule
    $N_{\text{pm}}$ & 1 \\ \midrule
    Size of $\text{ArgNet}$'s output & 5 \\ \midrule
    Learning rate (LR) for $\mathcal{L}_{\text{MLM}}$ & 3e-4  \\ \midrule
    LR for $\mathcal{L}_{\text{VAE}}$ &   3e-4  \\ \midrule
    VAE encoder network & [256, 256, 32], relu \\ \midrule
    VAE decoder network & [256, 256, $|\tl|$], relu \\  \midrule
    Predicate network & [128, 128, 2], relu \\  \midrule
    Argument network & [128, 128, 5], relu \\ \midrule
    PPO epoch & 10 \\ \midrule
    PPO policy LR & 3e-4 \\ \midrule
    PPO value LR & 3e-4 \\ \midrule
    PPO policy network & [32, 64, 64], tanh \\ \midrule
    PPO value network &  [32, 64, 64], tanh \\ \midrule
    PPO mini-batch size & 128 \\ \midrule
    PPO nums of mini-batch & 160 \\ 
    \bottomrule
    \end{tabular}
    
    \label{tab:hyper-parameters}
\end{table}

\clearpage
\section{Additional Experiment Results}
\label{appendix:extra_experiment}

\subsection{Training IFP with Different Number of NL Sentence Patterns}
We evaluate the robustness of different methods by training an IFP on the tasks with 1, 5, and 9 NL sentence patterns for 2M timesteps. Table \ref{tab:different_num_of_pattern} presents the experiment results. As the number of sentence patterns increases, \methodname~nearly maintains a near 100\% success rate. Since \methodname~could uniquely represent the NL expressions in different sentence patterns, the IFP policy that follows TL can comprehend the instructions more easily and learn to manage the task rapidly. This result indicates that learning a unique representation of human instruction benefits policy learning. In contrast, as the number of NL sentence patterns increases, the performance of baselines declines significantly because they must simultaneously comprehend more NL expressions and learn the skills to manage the task.

\begin{table}[htbp]
    \centering
    \caption{A summary of the final success rate (\%) on the training set with a different number of NL sentence patterns. Each IFP is trained for 2M timesteps and evaluated for 40 episodes. The results are averaged over $5$ seeds.}
    \begin{tabular}{c|c|c|c}
    \toprule
    \diagbox{Method}{Pattern nums.}  & 1 & 5 & 9 \\   
    \midrule
    \methodname & \textbf{100\%} & \textbf{100\%} & \textbf{99.9\%}  \\ \midrule
    One-hot & $98.2\%$ & $91.8\%$  & $86.5\%$ \\ \midrule
    Bert-binary & $82.0\%$  & $60.9\%$ & $64.0\%$ \\ \midrule
    Bert-continuous & $95.3\%$ & $63.1\%$ & $60.7\%$ \\ 
    \bottomrule
    \end{tabular}
    \label{tab:different_num_of_pattern}
\end{table}

\subsection{Complete Results of T-SNE Projection of Different Representations}
\label{appendix:extra_exp_tsne}

\begin{figure}[htbp]
\centering
\subcaptionbox*{}{
\rotatebox{90}{\ \ \ \ \ \ \ \ \ \ \ \ \ \large  TL-predicate}
\includegraphics[width=0.25\textwidth]{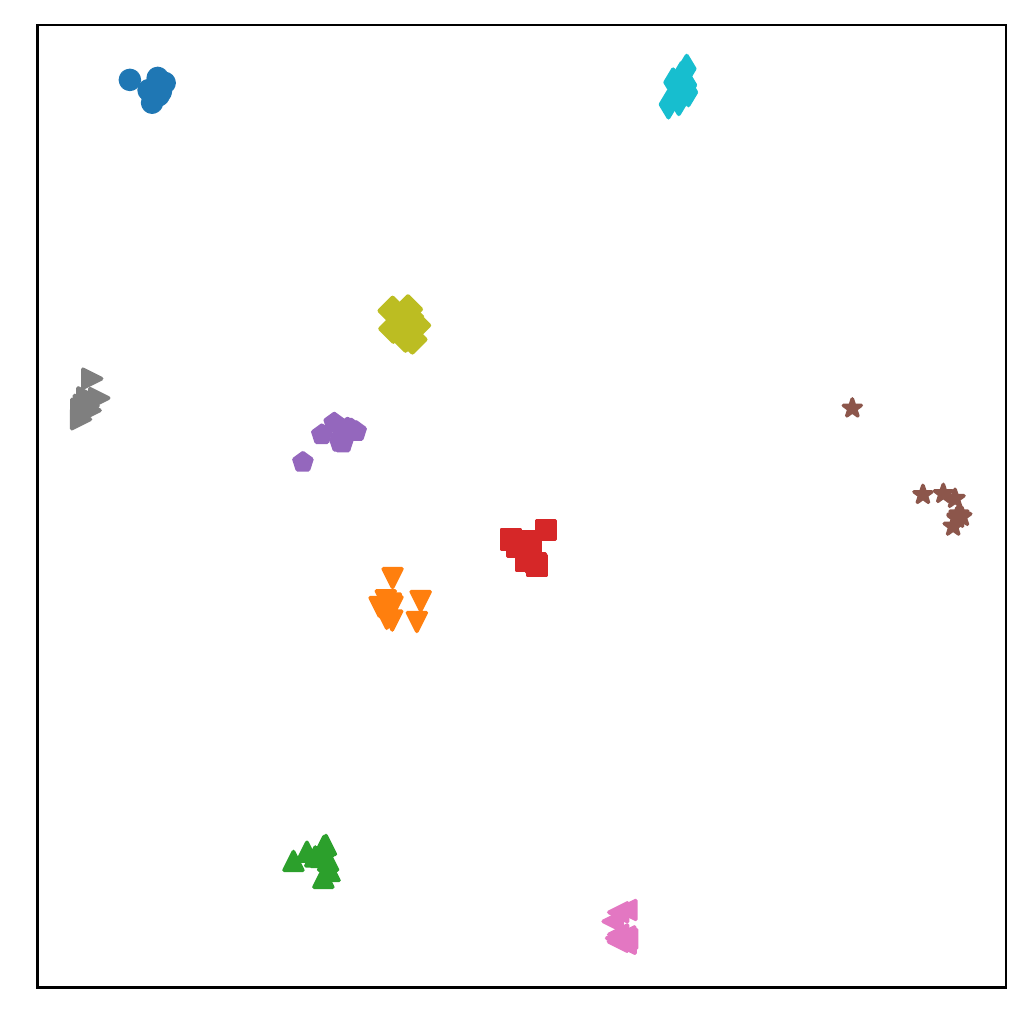}
}
\hspace{1em}
\subcaptionbox*{}{
\includegraphics[width=0.25\textwidth]{tsne/translator_our_test.pdf}
}
\hspace{1em}
\subcaptionbox*{}{
\includegraphics[width=0.25\textwidth]{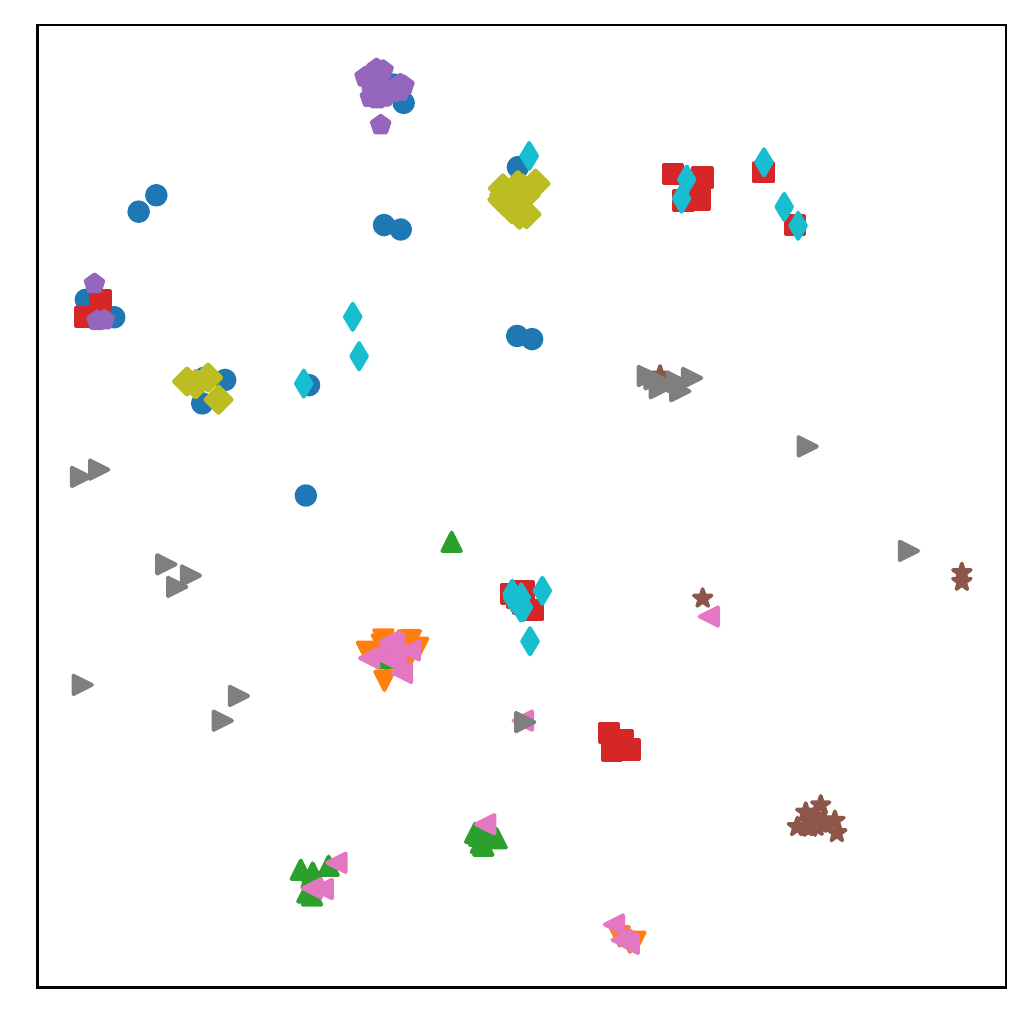}
}\\
\vspace{-1.5em}
\subcaptionbox*{}{
\rotatebox{90}{\ \ \ \ \ \ \ \ \ \ \ \  \ \ \large  TL-binary}
\includegraphics[width=0.25\textwidth]{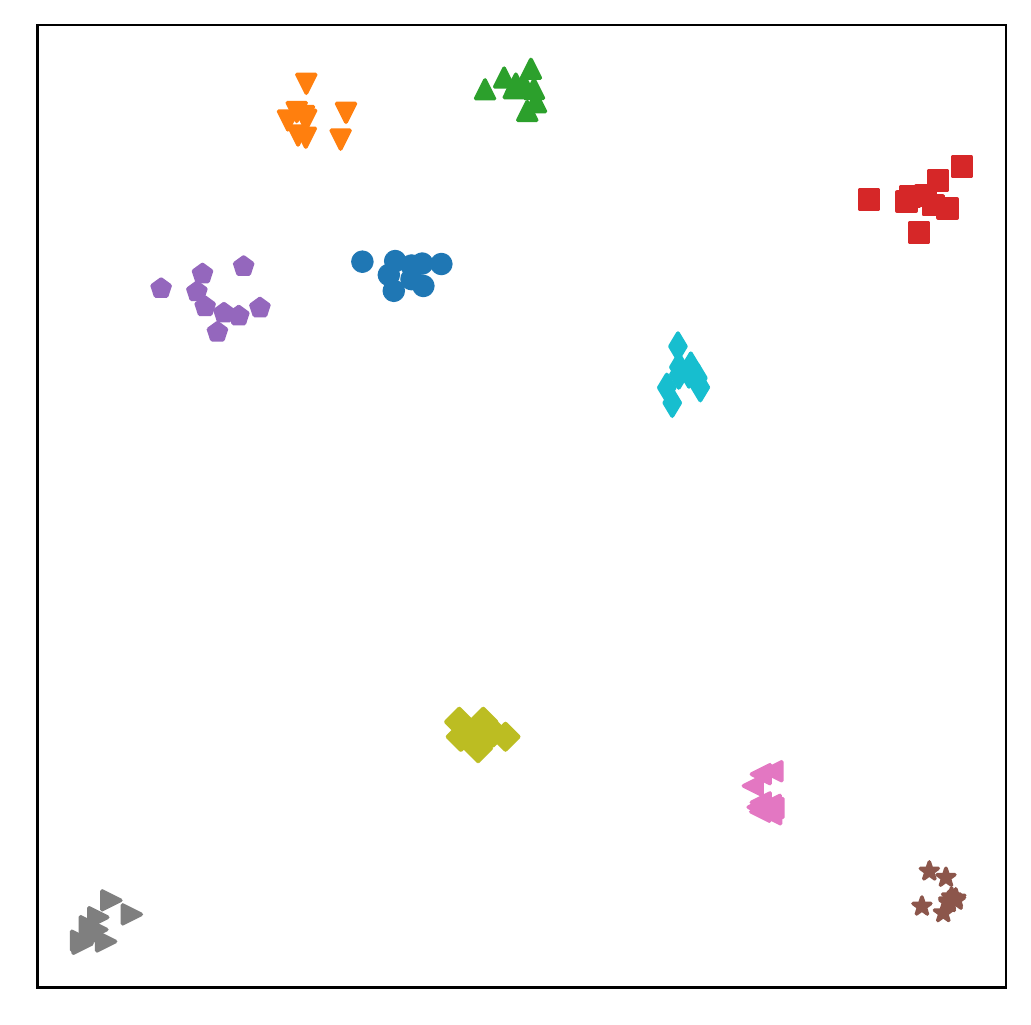}
}
\hspace{1em}
\subcaptionbox*{}{
\includegraphics[width=0.25\textwidth]{tsne/translator_binary_test.pdf}
}
\hspace{1em}
\subcaptionbox*{}{
\includegraphics[width=0.25\textwidth]{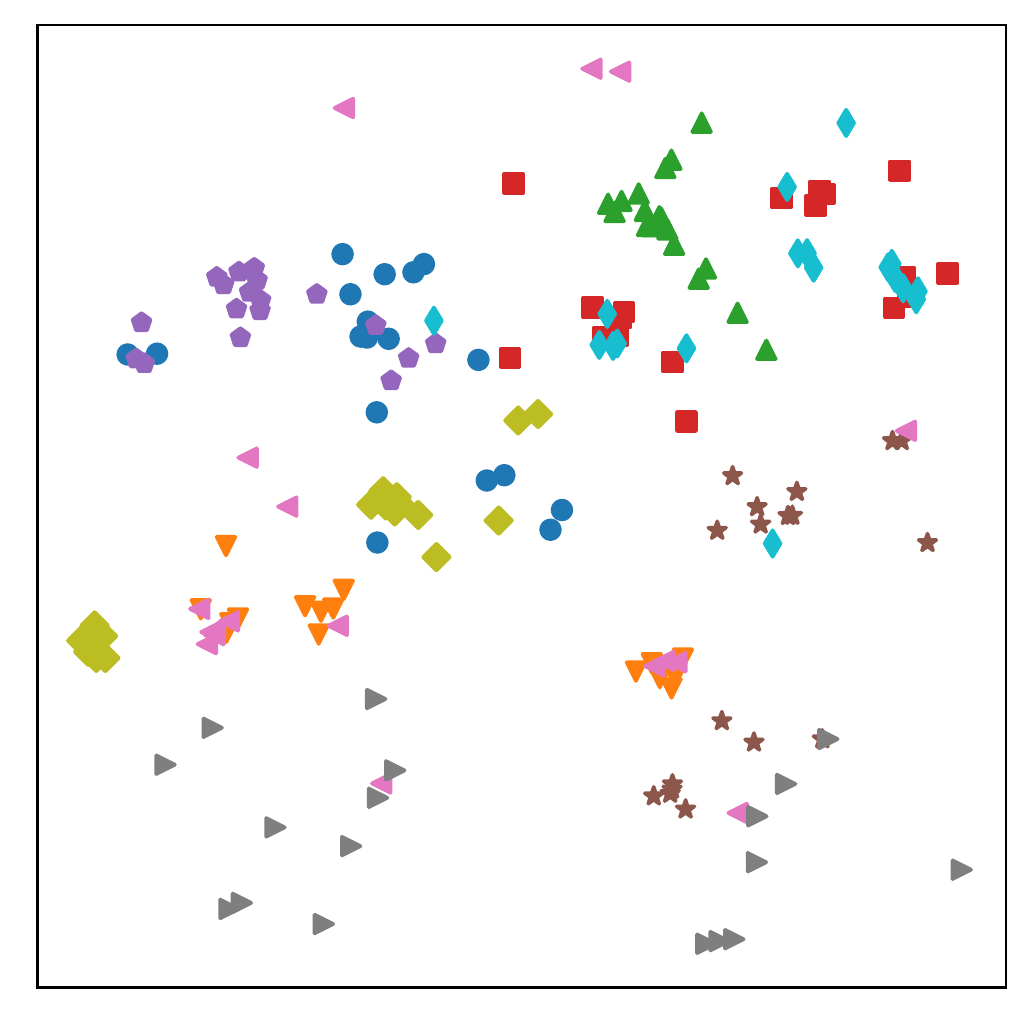}
}\\
\vspace{-1.5em}
\subcaptionbox*{}{
\rotatebox{90}{\ \ \ \ \ \  \ \ \ \ \ \large TL-continuous}
\includegraphics[width=0.25\textwidth]{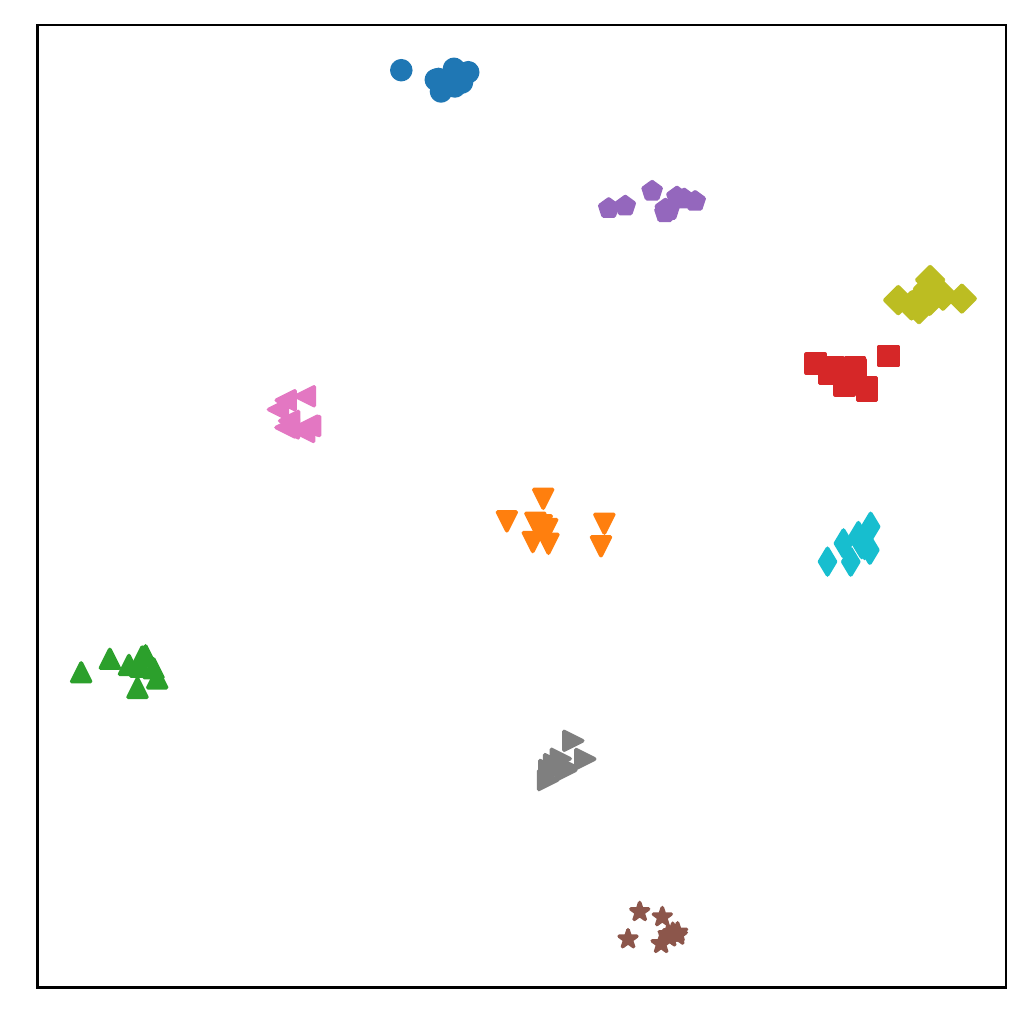}
}
\hspace{1em}
\subcaptionbox*{}{
\includegraphics[width=0.25\textwidth]{tsne/translator_cont_test.pdf}
}
\hspace{1em}
\subcaptionbox*{}{
\includegraphics[width=0.25\textwidth]{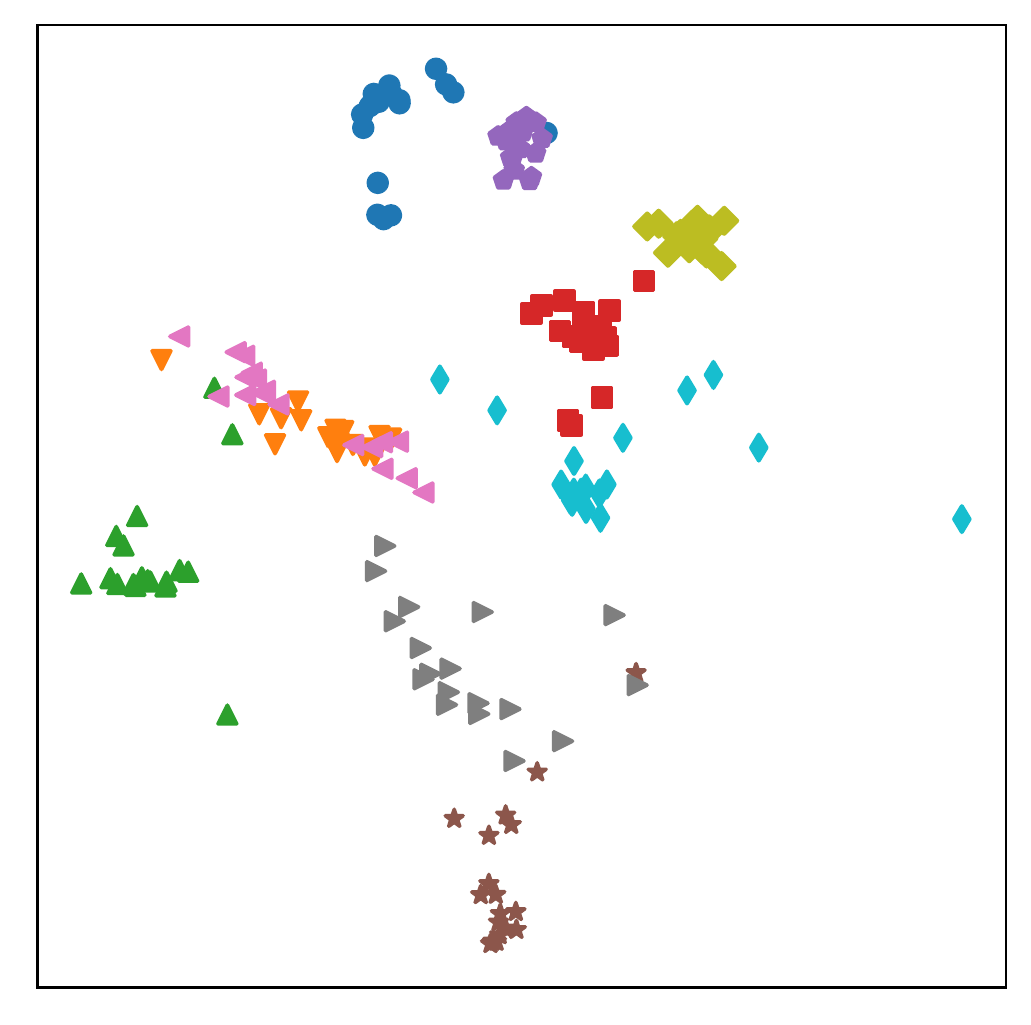}
}\\
\vspace{-1.5em}
\subcaptionbox*{}{
\rotatebox{90}{\ \ \ \ \ \ \ \ \ \  \large Bert encoding}
\includegraphics[width=0.25\textwidth]{tsne/translator_bert_train.pdf}
}
\hspace{1em}
\subcaptionbox*{}{
\includegraphics[width=0.25\textwidth]{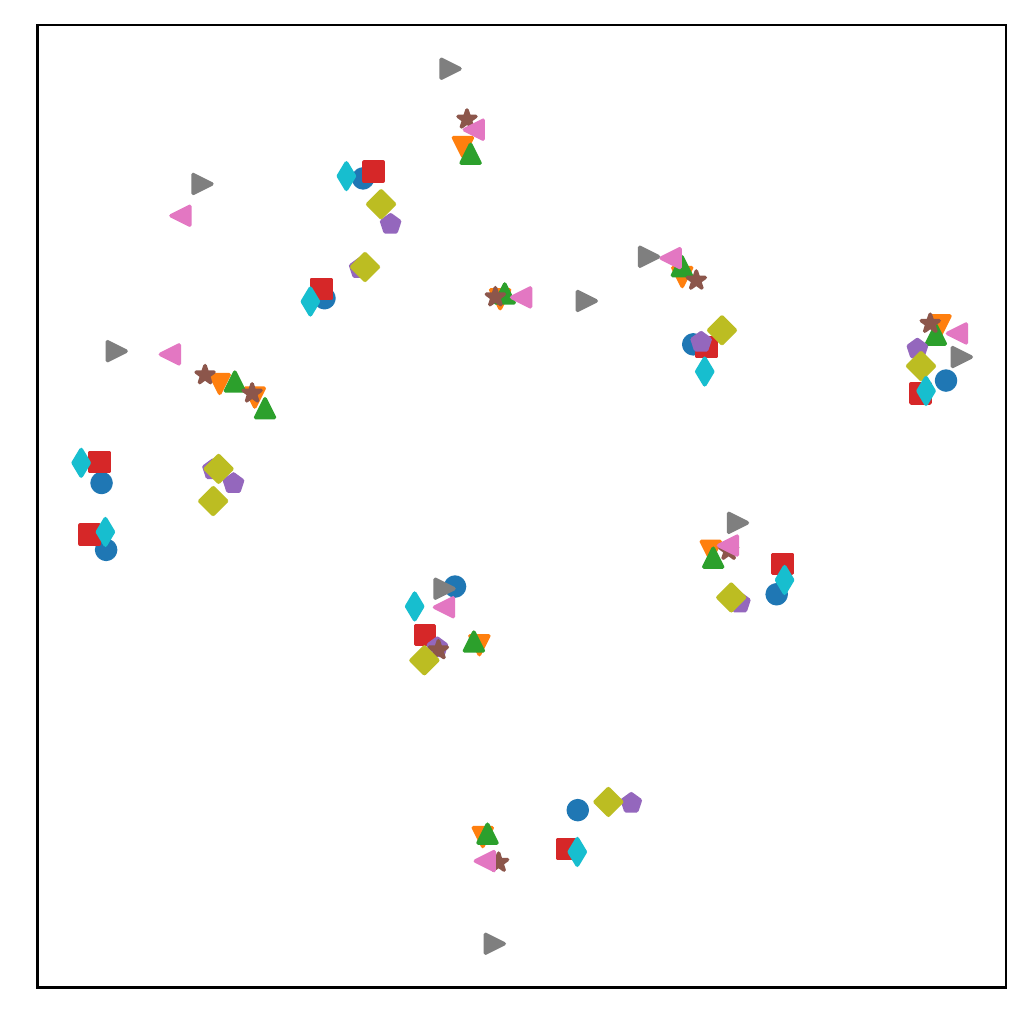}
}
\hspace{1em}
\subcaptionbox*{}{
\includegraphics[width=0.25\textwidth]{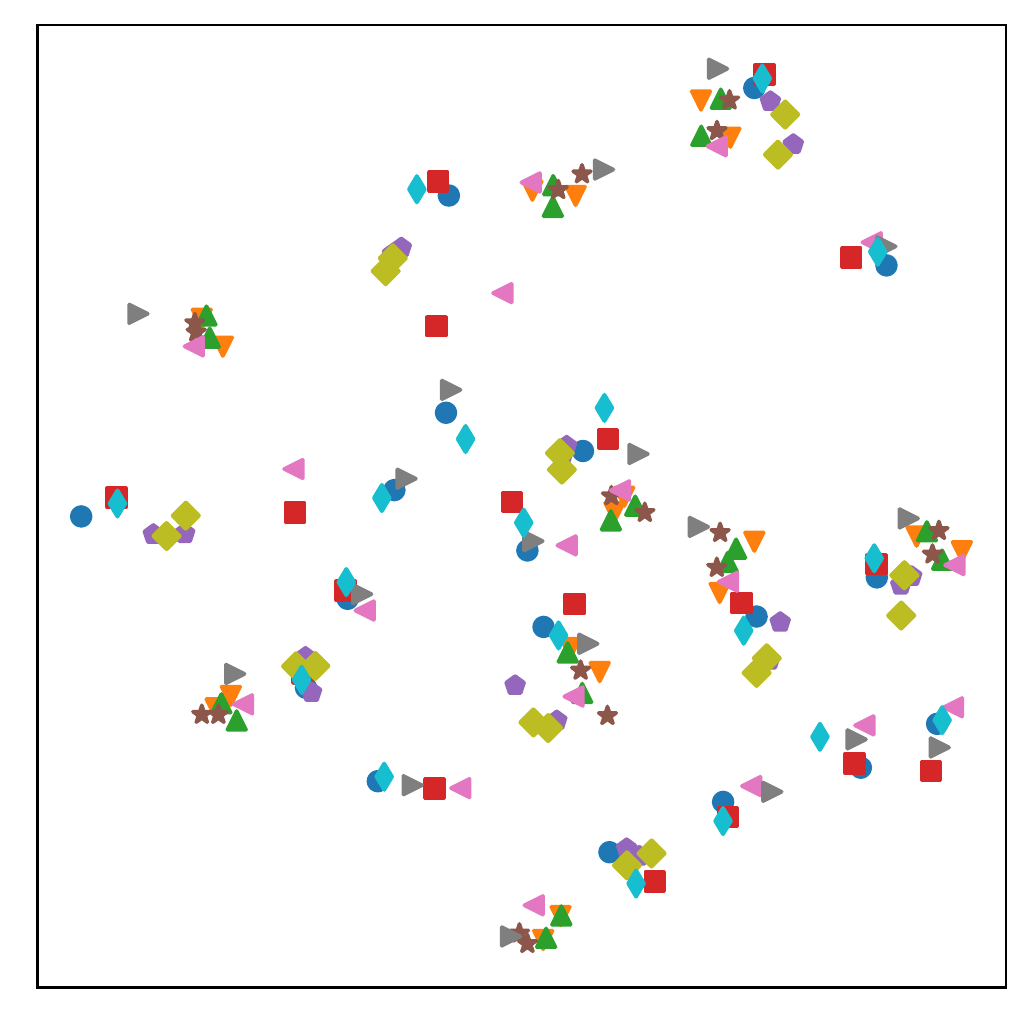}
}\\
\vspace{-1.5em}
\subcaptionbox*{\large Training}{
\rotatebox{90}{\ \ \ \ \ \ \ \  \large Bert-continuous}
\includegraphics[width=0.25\textwidth]{tsne/translator_bert_extractor_train.pdf}
}
\hspace{1em}
\subcaptionbox*{\large Testing}{
\includegraphics[width=0.25\textwidth]{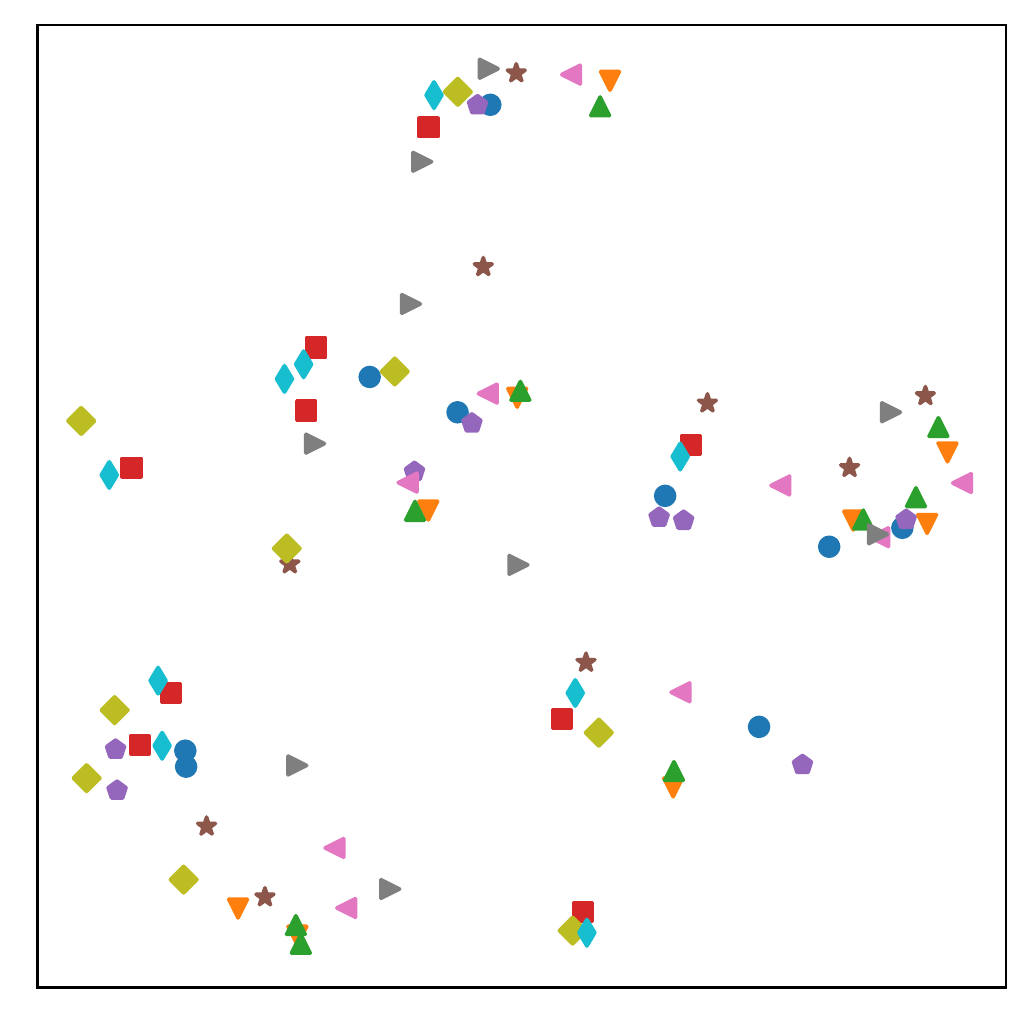}
}
\hspace{1em}
\subcaptionbox*{\large Error-added}{
\includegraphics[width=0.25\textwidth]{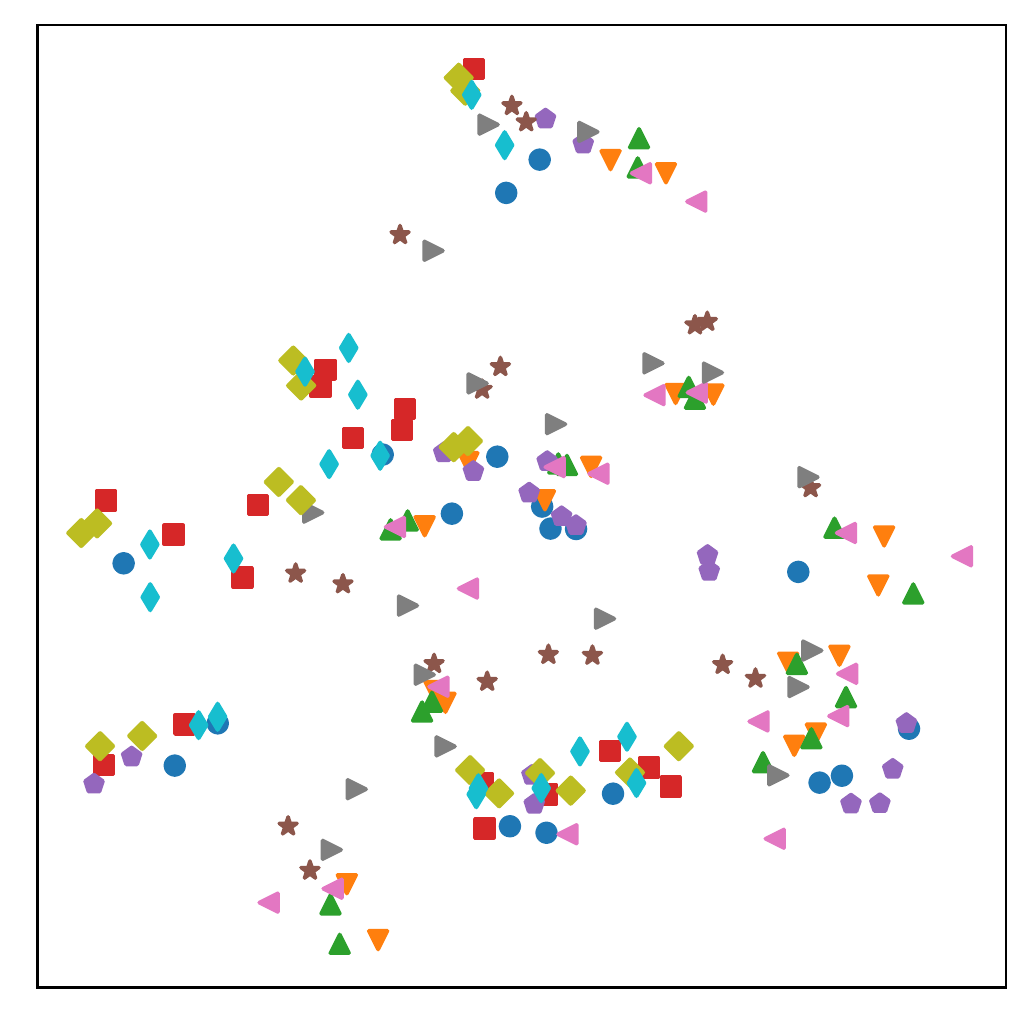}
}
\caption{The t-SNE projection of different representations on three NL expressions datasets. Each row represents one kind of representation, and each column stands for one kind of NL expression. Points with the same marker encode nine different NL expressions that describe the same human instruction.} 
\label{fig:full_tsne}
\end{figure}

Figure \ref{fig:full_tsne} presents the total experimental results of the t-SNE projection of various representations on three NL instruction datasets. \textbf{TL-predicate, TL-continuous and TL-binary} stand for TL with predicate, continuous, and binary representations. \textbf{Bert-encoding} is the output of the Bert model. \textbf{Bert-continuous} is the output of the OIL baseline's NL feature layer. We observe that, on the training NL instruction set, the t-SNE projections of the three TL representations are considerably more agminated than those of Bert encoding and Bert-continuous, indicating that IOL is an effective method for learning a unique representation. However, in the testing and error-added sets, the TL-predicate representation performs significantly better than TL-binary and TL-continuous, as its points with same marker are more concentrated. This result demonstrates that predicate representation is more resistant to unseen NL expressions and produces a more concentrated representation. In contrast, the t-SNE projections of Bert encoding and Bert-continuous are dispersed throughout the plane, inducing NL understanding burden for the policy learning.

\clearpage
\subsection{Deployment Examples of IFP Trained by \methodname}
\begin{figure}[htbp]
\centering
\subcaptionbox*{}{
\includegraphics[width=0.69 \textwidth]{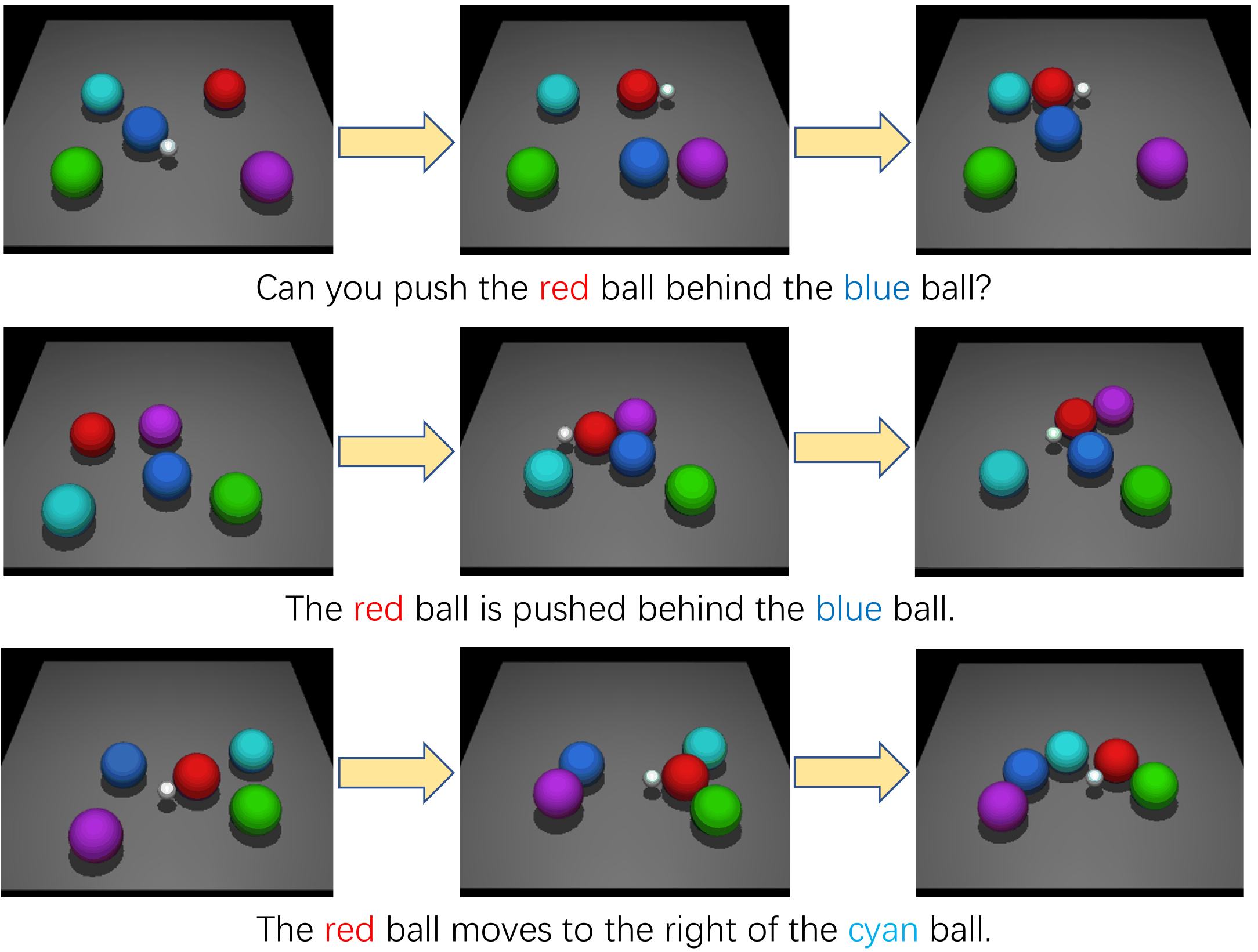}
}
\subcaptionbox*{}{
\includegraphics[width=0.69 \textwidth]{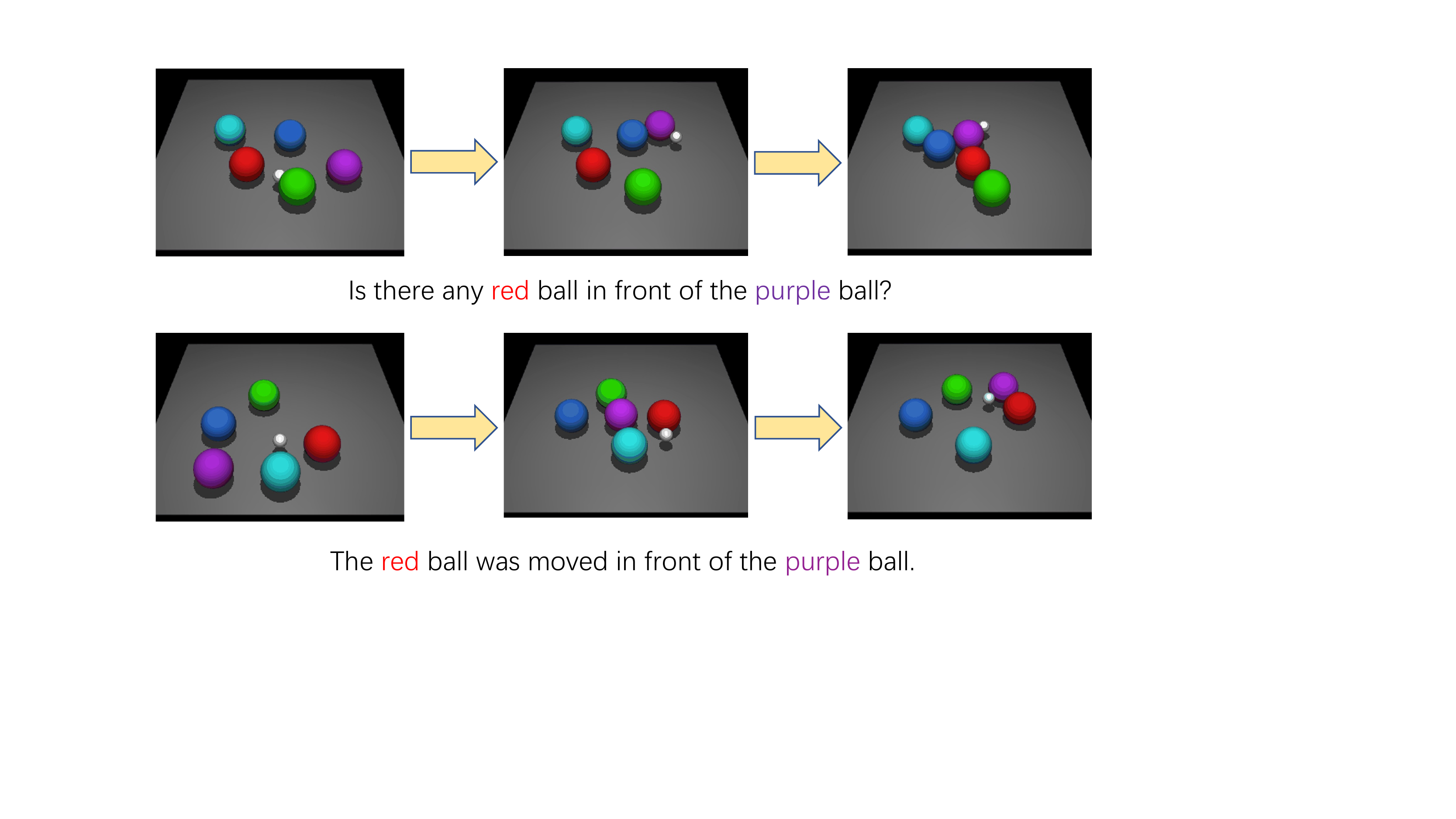}
}
\caption{A visualization of the \methodname's IFP deployment process.} 
\label{fig:ifp_example}
\end{figure}

We visualize the performance of the IFP trained by \methodname~in the CLEVR-Robot environment. Figure \ref{fig:ifp_example} presents the policy performance following different NL instructions. IFP can rapidly complete the tasks by moving related balls for NL instructions with different NL expressions.

\newpage
\subsection{Performance of \methodname~With Different Number of the Argument Networks}
\label{appendix:diff_num_of_arg_net}
\begin{figure}[htbp]
\centering
\subcaptionbox*{}{
\includegraphics[width=0.315\textwidth]{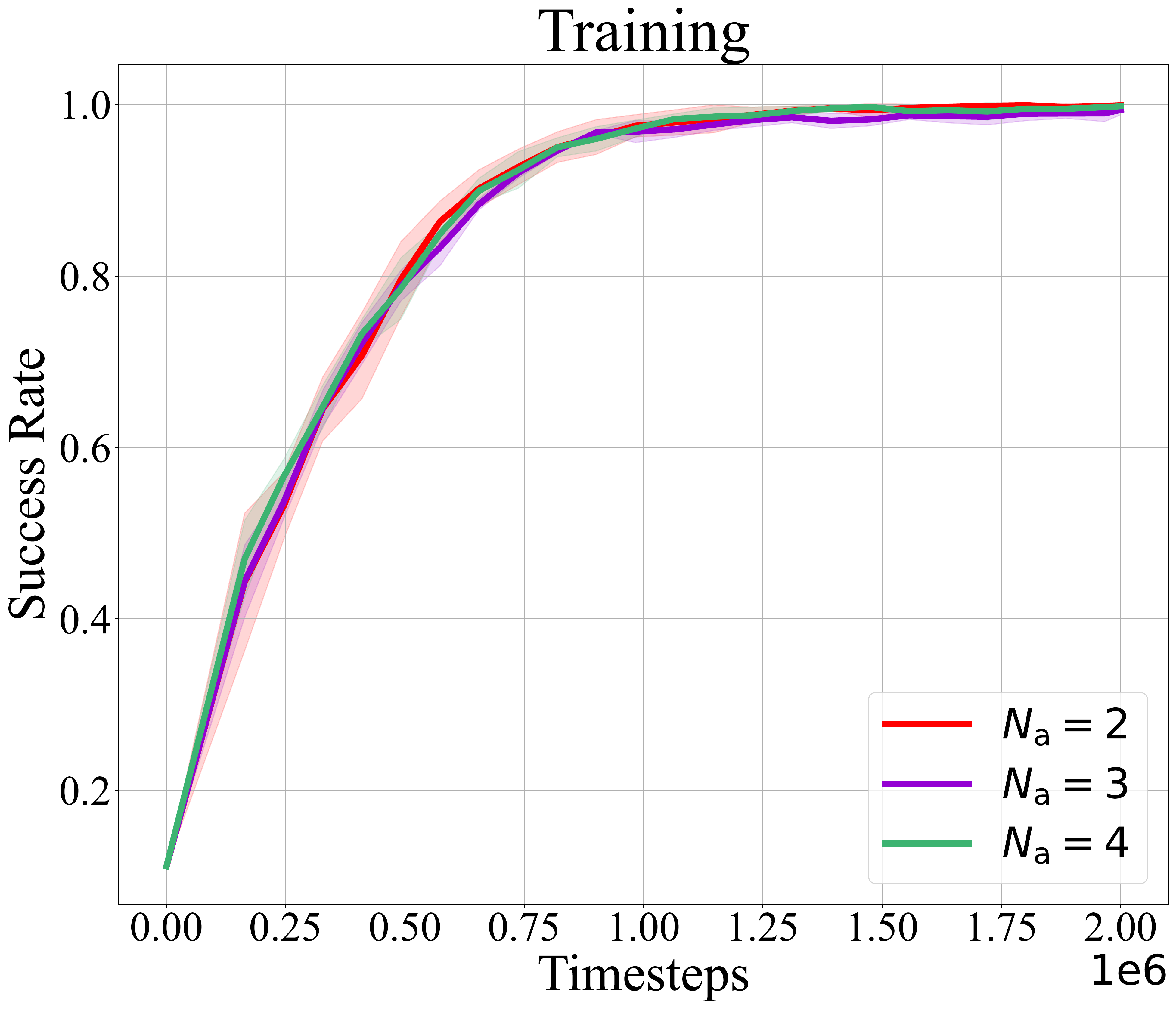}
}
\subcaptionbox*{}{
\includegraphics[width=0.315\textwidth]{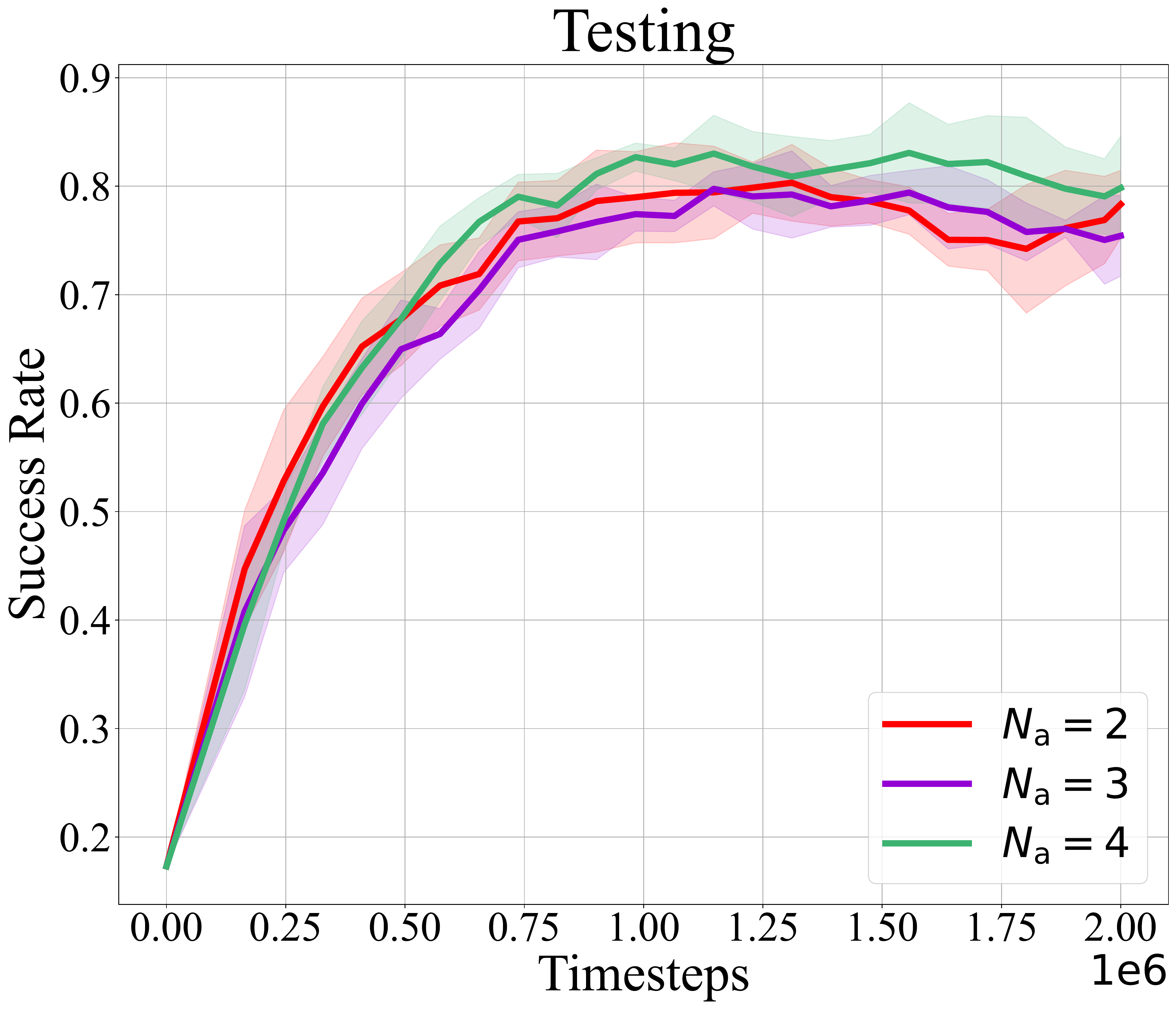}
}
\subcaptionbox*{}{
\includegraphics[width=0.315\textwidth]{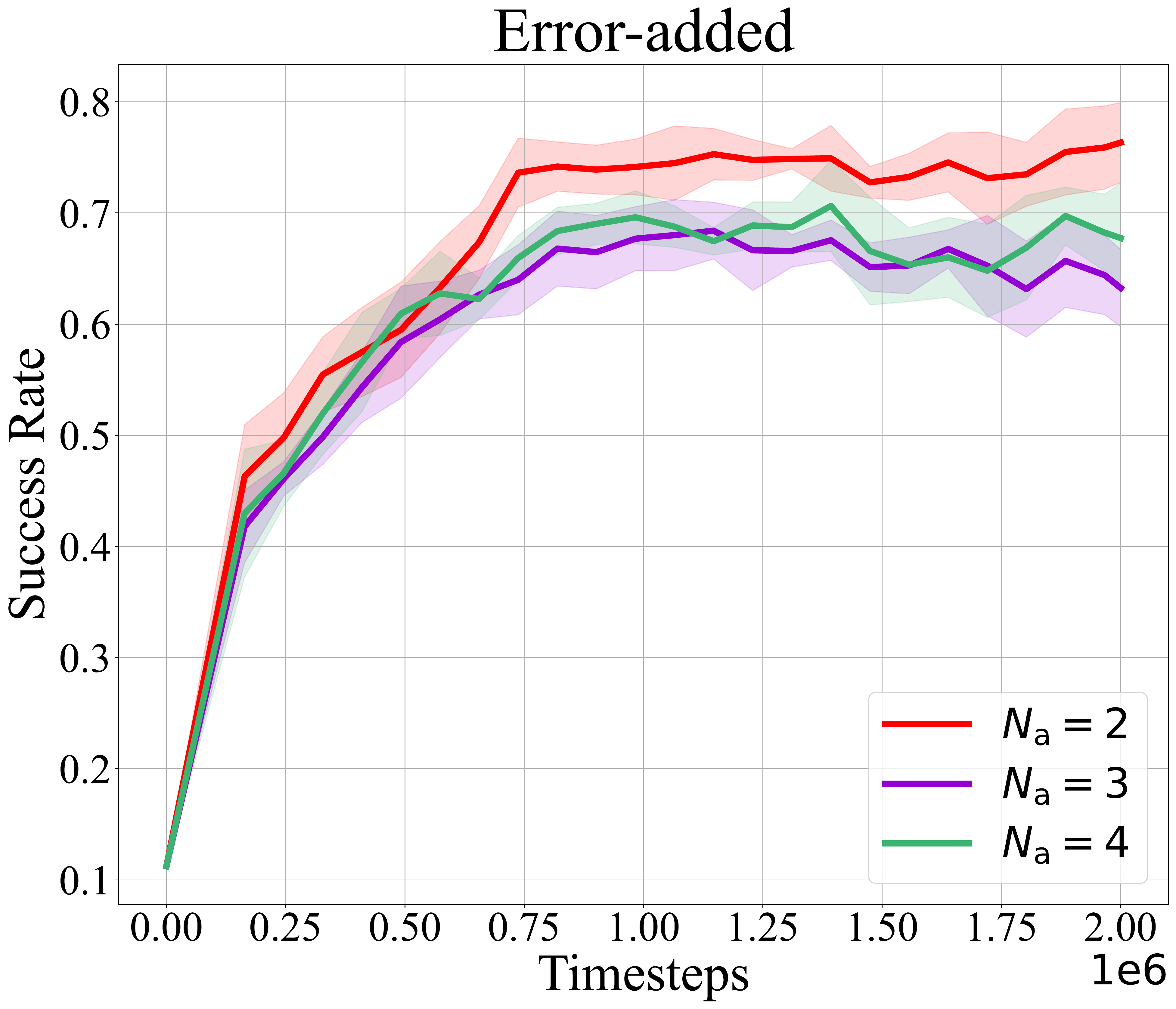}
}
\caption{Training curves of \methodname~with different number of argument networks. The x-axis represents the total timesteps agent interacts with the environment, and the y-axis represents the success rate of completing instructions. The shaded area stands for the standard deviation over five random trials.}
\label{fig:curve_diff_an}
\end{figure}

Figure \ref{fig:curve_diff_an} shows the training curves of \methodname~with different number of argument networks, when $N_{\text{pm}}=1$ and $N_{\text{pn}}=4$. The experiment results indicate that \methodname~is not sensitive to $N_{\text{a}}$, and \methodname~with $N_{\text{a}}=2$ performs slightly better than the other two parameters. 

\subsection{Training Curves of \methodname-MLP}
\label{appendix:extra_mlp}

\begin{figure*}[htbp]
\centering
\subcaptionbox*{}{
\includegraphics[width=0.315\textwidth]{exp_fig/mlp_train.pdf}
}
\subcaptionbox*{}{
\includegraphics[width=0.315\textwidth]{exp_fig/mlp_test.pdf}
}
\subcaptionbox*{}{
\includegraphics[width=0.315\textwidth]{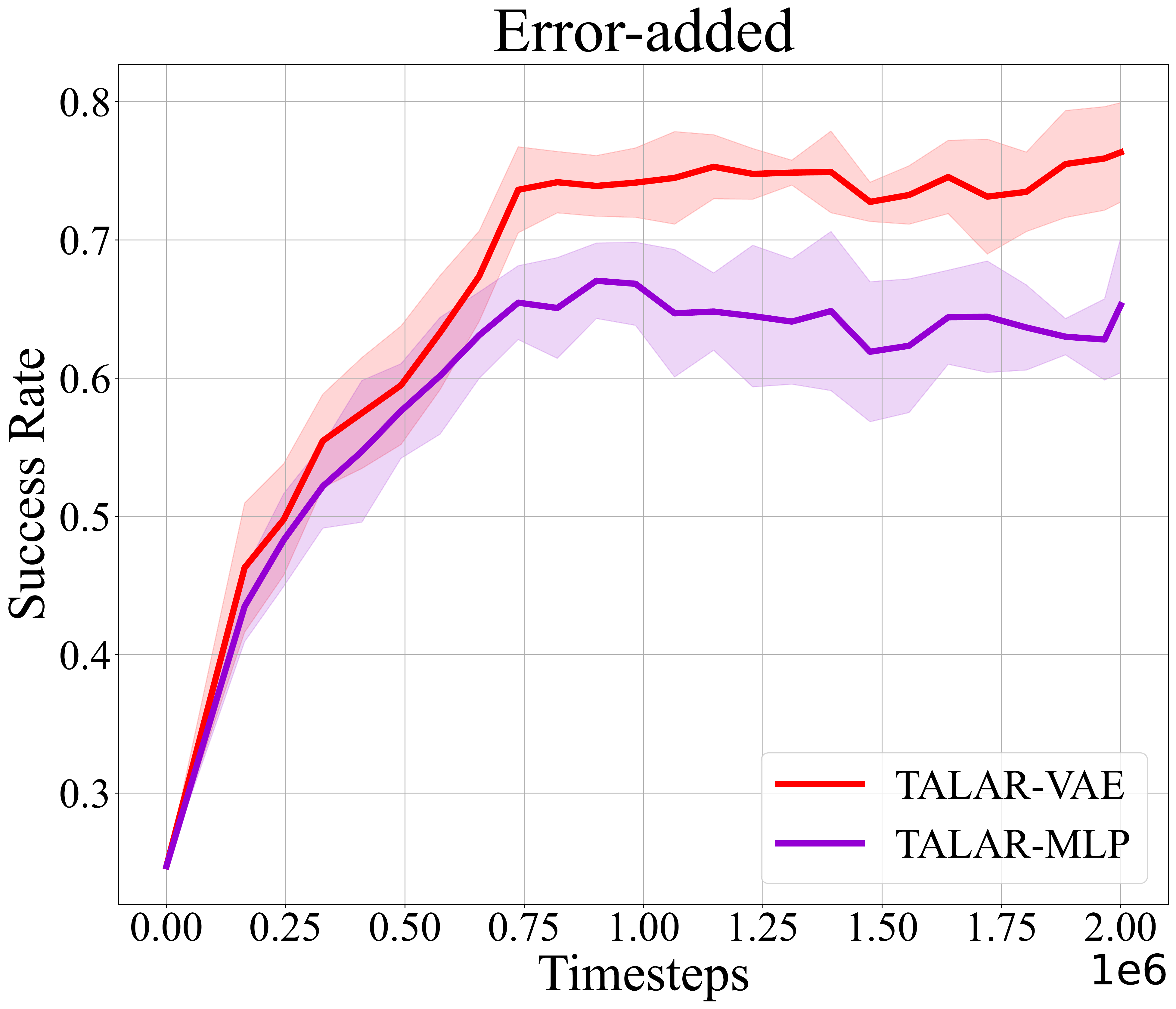}
}
\caption{Ablation study on the VAE used in \methodname. The x-axis represents the total timesteps agent interacts with the environment, and the y-axis represents the success rate of completing instructions. The shaded area stands for the standard deviation over five random trials.}
\label{fig:mlp_results}
\vspace{-1em}
\end{figure*}

Figure \ref{fig:mlp_results} shows the ablation study on the VAE used in \methodname, where \methodname-MLP replaces the VAE in \methodname~with an MLP network and trains the translator with a supervised learning loss. The experimental results indicate that \methodname-VAE is more robust on unseen NL instructions than \methodname-MLP, while they achieve similar performance on the training dataset.

\clearpage
\section{Retrievals for Notations and Abbreviations}

\begin{table}[htbp]
    \centering
    \caption{Notations and abbreviations in this paper.}
    \begin{tabular}{cc}
    \toprule
    \textbf{Name} & \textbf{Meaning}  \\   \toprule
    \textbf{Notations} \\ 
    $\tl$ & task language \\
    $\nl$ & natural language \\
    $q_{\phi_1}$ & encoder with parameters $\phi_1$ \\
    $p_{\phi_2}$ & decoder with parameters $\phi_2$ \\
    $g_\theta$ & TL generator with parameters $\theta$ \\ \toprule
    \textbf{Abbreviations} \\
    NL & Natural Language  \\
    TL & Task Language  \\
    LM & Language Model \\
    RL & Reinforcement Learning \\
    PM & Predicate Module  \\
    IOL & Inside-Out Learning  \\ 
    OIL & Outside-In Learning  \\ 
    VAE & Variational Auto-Encoder \\ 
    MLM & Masked Language Modelling \\ 
    NLC-RL & Natural Language-Conditioned Reinforcement Learning  \\ 
    \bottomrule
    \end{tabular}
    \label{tab:notations}
\end{table}

\end{document}